\documentclass[accepted]{uai2021} 

\usepackage[american]{babel}

\usepackage{natbib} 
\usepackage{mathtools} 
\usepackage{booktabs} 
\usepackage{tikz} 
\usepackage{amsmath, amssymb}
\usepackage{subcaption}
\usepackage[capitalize]{cleveref}
\usepackage{pgfplots, pgfplotstable}
\usetikzlibrary{pgfplots.groupplots}
\pgfplotsset{compat=1.16}
\usetikzlibrary{shapes.arrows, fadings, shapes}
\usetikzlibrary{arrows}
\usetikzlibrary{positioning}
\usetikzlibrary{patterns}

\definecolor{nice-red}{HTML}{E41A1C}
\definecolor{nice-orange}{HTML}{FF7F00}
\definecolor{nice-yellow}{HTML}{FFC020}
\definecolor{nice-green}{HTML}{4DAF4A}
\definecolor{nice-blue}{HTML}{377EB8}
\definecolor{nice-purple}{HTML}{984EA3}
\definecolor{nice-grey}{HTML}{6C7A89}
\definecolor{nice-pink}{HTML}{DB5A6B}

\newcommand\argmax{\ensuremath{\arg\max}}
\DeclareMathAlphabet{\mathcal}{OMS}{cmsy}{m}{n}

\title{Learning Emergent Discrete Message Communication \\ for Cooperative Reinforcement Learning}

\author[1]{Sheng~Li} 
\author[2]{Yutai~Zhou}
\author[2]{Ross~Allen}
\author[1]{Mykel~J.~Kochenderfer}
\affil[1]{%
    Aero/Astro Engineering\\
    Stanford University\\
    Stanford, CA, USA
}
\affil[2]{%
    MIT Lincoln Lab\\
    Lexington, MA, USA\\
}

\begin{document}
\maketitle


\begin{abstract}
Communication is a important factor that enables agents work cooperatively in multi-agent reinforcement learning (MARL). Most previous work uses continuous message communication whose high representational capacity comes at the expense of interpretability. Allowing agents to learn their own discrete message communication protocol emerged from a variety of domains can increase the interpretability for human designers and other agents.This paper proposes a method to generate discrete messages analogous to human languages, and achieve communication by a broadcast-and-listen mechanism based on self-attention. We show that discrete message communication has performance comparable to continuous message communication but with much a much smaller vocabulary size.Furthermore, we propose an approach that allows humans to interactively send discrete messages to agents. 
\end{abstract}
\section{Introduction}\label{sec:intro}

Communication allows agents to share information  so that they can perform tasks cooperatively. There has been existing work on using deep reinforcement learning (RL) to produce communication protocols. For example, CommNet~\citep{sukhbaatar2016learning}, a recurrent communication model, averages the hidden states for centralized communication. 
IC3Net~\citep{singh2018learning}, an extension on CommNet, adopts a more complicated but similar centralized aggregation approach to communication.
Instead of centralized aggregation and averaging, TarMAC~\citep{das2019tarmac} use multi-headed attention to distribute information to other agents.
BiCNet~\citep{peng2017multiagent} and ATOC~\citep{jiang2018learning} both use a bidirectional recurrent network as a communication channel. They fix the positions of agents in the bidirectional recurrent network to specify their roles. 
DICG~\citep{li2020deep} uses graph convolution to implicitly pass information between agents.

However, typically, existing multi-agent communication approaches use continuous messages to communicate. They use real-valued vectors to encode messages. Human languages, however, use discrete characters and words. An advantage of continuous messaging is its representational capacity, but it can be at the expense of interpretability from the perspective of human designers or other agents.

We propose a deep RL model for agents to learn to generate their own discrete message protocols. Our model produces discrete messages by identifying the maximum element in message vectors, resulting in greater stability than sampling. The model adopts a broadcast-and-listen procedure to send and receive messages. It uses self-attention mechanism~\citep{cheng2016long} to aggregate messages sent by other agents. The model is differentiable and therefore can be learned end-to-end. \citet{evtimova2017emergent} use bit-string messaging to learn emergent communication in referential games for two agents. Our approach is applicable to any number of agents.

We compare the performance of discrete message communication with continuous message communication in a variety of domains, showing that discrete message communication has comparable performance to continuous message communication with a much smaller vocabulary size. We also study the effects of communication bandwidth and vocabulary size on discrete message communication, using the metrics positive listening and positive signaling, where positive listening indicates received messages are influencing agents’ behaviors in some way, and positive signaling indicates an agent is sending messages that are related in some way with its own observations or actions~\citep{lowe2019pitfalls, jaques2019social}.
Furthermore, we propose an approach for human-agent interaction using discrete message communication, demonstrating its interpretability.
\section{Background}\label{sec:bg}
We represent the problem as a Dec-POMDP~\citep{oliehoek2016concise} defined by the tuple $\langle \mathcal{I}, \mathcal{S}, \{\mathcal{A}^i\}_{i=1}^n, \mathcal{V}, \mathcal{T}, \mathcal{Z}, R, \mathcal{O}, \gamma\rangle$, where $\mathcal{I} = \{1, \ldots, n\}$ is the set of agents, $\mathcal{S}$ is the global state space, $\mathcal{A}^i$ is the action space of the $i$th agent, and $\mathcal{Z}$ is the observation space for an agent.
The discrete communication vocabulary set is defined by $\mathcal{V} = \{0, 1\}^b$, where $b$ is the band width of communication. A message from $\mathcal{V}$ is therefore a binary vector.
The transition function defining the next state distribution is given by $\mathcal{T}: \mathcal{S} \times \prod_i \mathcal{A}^i \times \mathcal{S} \to [0,1]$.
The reward function is $R: \mathcal{S} \times \prod_i \mathcal{A}^i \to \mathbb{R}$, and the discount factor is $\gamma \in [0, 1)$.
The observation model defining the observation distribution from the current state is $\mathcal{O}: \mathcal{S} \times \mathcal{Z} \to [0, 1]$.
Each agent $i$ has a stochastic policy $\pi^i$ conditioned on its observations $o_i$.
The discounted return is $G_t = \sum_{l=0}^\infty \gamma^{l} r_{t+l}$, where $r_t$ is the joint reward at step $t$.
The joint policy $\mathbf{\pi}$ induces a value function $V^{\mathbf{\pi}}(s_t) = \mathbb{E}[G_t \mid s_t]$ and an action-value function $Q^\mathbf{\pi}(s_t, \mathbf{a}_t) = \mathbb{E}[G_t \mid s_t, \mathbf{a}_t]$, where $\mathbf{a}_t$ is the joint action.
The advantage function is $A^\mathbf{\pi}(s_t, \mathbf{a}_t)=Q^\mathbf{\pi}(s_t, \mathbf{a}_t) - V^{\mathbf{\pi}}(s_t)$.

\subsection{Policy Optimization}\label{sec:ppo}
We use policy optimization to maximize the expected discounted return.
Given policy $\pi_\theta$ parameterized by $\theta$, the surrogate policy optimization objective is~\citep{schulman2017proximal}:
\begin{equation}
    \underset{\theta}{\mathrm{maximize}} \quad \hat{\mathbb{E}}_t\left[\frac{\pi_{\theta}(a_t \mid s_t)}{\pi_{\theta_{\text{old}}}(a_t \mid s_t)} \hat{A}_t\right],
\end{equation}
where we use the generalized advantage estimation (GAE)~\citep{schulman2015high} to estimate $\hat{A}_t$ at time step $t$, and the expectation $\hat{\mathbb{E}}_t[\cdot]$ indicates the empirical average over a finite batch of samples.
In practice, we use the clipped PPO objective \citep{schulman2017proximal} to limit the step size for stable updates. The entropy bonus is also added to encourage exploration. The objective to maximize becomes
\begin{equation}\label{eq:ppo_obj}
\small
\begin{aligned}
    L^{\mathrm{PPO}}_t(\theta) = \hat{\mathbb{E}}_t \Big[&\min\big(r_t(\theta) \hat{A}_t,\,\, \mathrm{clip}(r_t(\theta), 1 - \epsilon, 1 + \epsilon)\hat{A}_t \big) \\
    &+ \beta H[\pi_\theta](s_t)\Big],
\end{aligned}
\end{equation}
where $r_t(\theta) = \frac{\pi_{\theta}(a_t \mid s_t)}{\pi_{\theta_{\text{old}}}(a_t \mid s_t)}$, and $H[\pi_\theta](s_t)$ is the entropy of the policy given state $s_t$, the clipping parameter $\epsilon$ and the entropy coefficient $\beta$ are hyperparameters.
In the context of centralized training but decentralized execution, we employ a \emph{parameter sharing} strategy whereby each agent in a homogeneous team uses identical copies of policy parameters~\citep{gupta2017cooperative}.

\subsection{Self-attention}
Self-attention mechanism~\citep{cheng2016long} emerged from the natural language processing community.
It is used to relate different positions of a single sequence. The difference between self-attention and standard attention is that self-attention uses a single sequence as both its source and target sequence. It has been shown to be useful in image caption generation~\citep{liu2018show, yu2019multimodal} and machine reading comprehension~\citep{cheng2016long, yu2018qanet}.

The attention mechanism has also been adopted recently for multi-agent reinforcement learning. The relations between a group of agents can be learned through attention. \citet{iqbal2018actor} use attention to extract relevant information of each agent from the other agents. \citet{jiang2018learning} use self-attention to learn when to communicate with neighboring agents. \citet{wright2019attentional} use self-attention on the policy level to differentiate different types of connections between agents. \citet{Jiang2020Graph} use multi-head dot product attention to compute interactions between neighboring agents for the purpose of enlarging agents' receptive fields and extracting latent features of observations. 
\citet{li2020deep} use multiplicative self-attention to implicitly build coordination graphs by weighting the graph edges with attention weights.

We use self-attention to learn the attention weights between agents. The attention weights are used to differentiate the importance of messages in the public communication channel for each agent.

\section{Approach}\label{sec:approach}
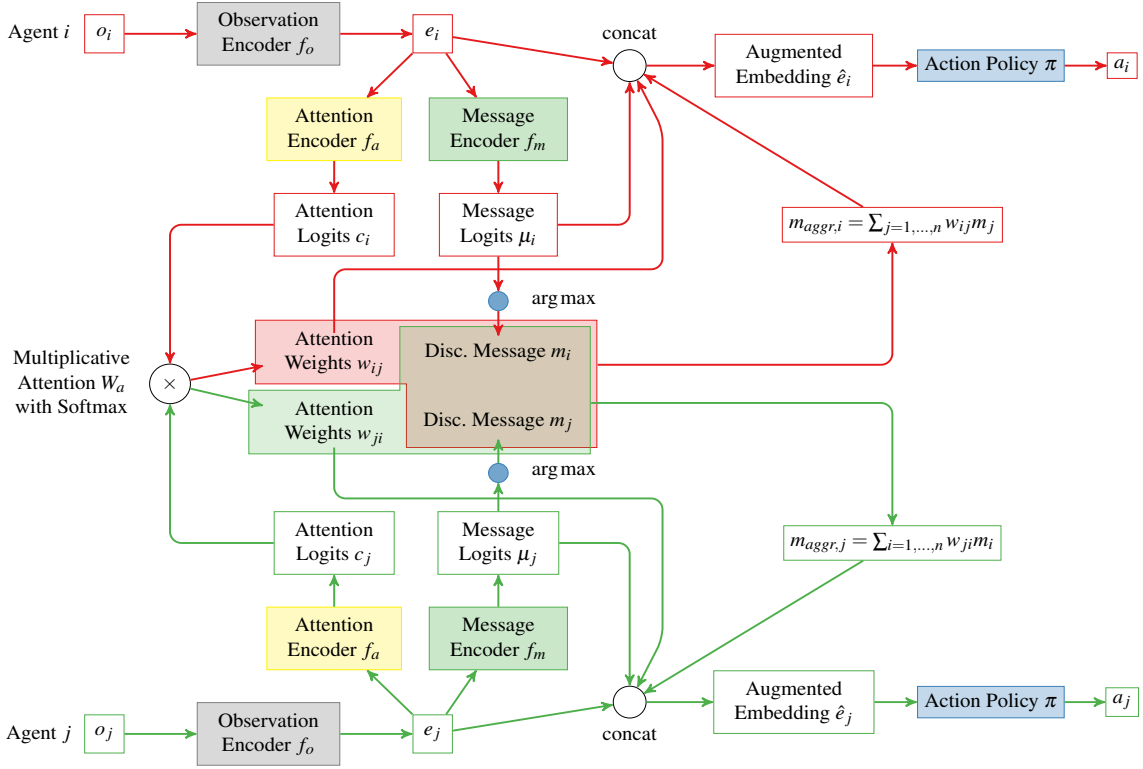
\begin{figure*}[t]
    \centering
    \resizebox{6in}{4in}{

\begin{tikzpicture}[>=stealth']
\small
\node[] () at (0, 0) {Agent $i$};
\node[rectangle, draw=nice-red, minimum height=0.6cm, minimum width=0.6cm,] (_o_i) at (1, 0) {$o_i$};
\node[rectangle, draw=gray, fill=gray!30] (_obs_encoder_i) at (3.5, 0) {\begin{tabular}{c}Observation \\ Encoder $f_o$ \end{tabular}};
\node[rectangle, draw=nice-red, minimum height=0.6cm, minimum width=0.6cm,] (_e_i) at (6, 0) {$e_i$};
\node[rectangle, draw=yellow, fill=yellow!30] (_attn_encoder_i) at (4.5, -1.5) {\begin{tabular}{c}Attention \\ Encoder $f_a$\end{tabular}};
\node[rectangle, draw=nice-green, fill=nice-green!30] (_msg_encoder_i) at (7, -1.5) {\begin{tabular}{c}Message \\ Encoder $f_m$\end{tabular}};

\draw[->, draw=nice-red, line width=0.3mm] (_o_i) to (_obs_encoder_i);
\draw[->, draw=nice-red, line width=0.3mm] (_obs_encoder_i) to (_e_i);
\draw[->, draw=nice-red, line width=0.3mm] (_e_i) to (_attn_encoder_i);
\draw[->, draw=nice-red, line width=0.3mm] (_e_i) to (_msg_encoder_i);

\node[rectangle, draw=nice-red] (_attn_logits_i) at (4.5, -3) {\begin{tabular}{c}Attention \\ Logits $c_i$ \end{tabular}};
\node[rectangle, draw=nice-red] (_msg_logits_i) at (7, -3) {\begin{tabular}{c}Message \\ Logits $\mu_i$ \end{tabular}};

\draw[->, draw=nice-red, line width=0.3mm] (_attn_encoder_i) to (_attn_logits_i);
\draw[->, draw=nice-red, line width=0.3mm] (_msg_encoder_i) to (_msg_logits_i);

\fill[draw=nice-red, fill=nice-red, fill opacity=0.2]
  (3.3, -4.5) -- (8.5, -4.5) -- (8.5, -6.5) -- (5.6, -6.5) -- (5.6, -5.5) -- (3.3, -5.5) -- cycle;
\node[] (_union_i) at (8.4, -5.2) {};
 
\fill[draw=nice-green, fill=nice-green, fill opacity=0.2]
  (3.2, -6.6) -- (8.4, -6.6) -- (8.4, -4.6) -- (5.5, -4.6) -- (5.5, -5.6) -- (3.2, -5.6) -- cycle;
\node[] (_union_j) at (8.3, -5.8) {};

\node[circle, draw=nice-blue, fill=nice-blue!70] (_argmax_i) at (7, -4.2) {};
\node[] () at (8, -4.2) {$\arg \max$};
\node[] (_disc_msg_i) at (7, -5) {Disc. Message $m_i$};

\draw[->, draw=nice-red, line width=0.3mm] (_msg_logits_i) to (_argmax_i);
\draw[->, draw=nice-red, line width=0.3mm] (_argmax_i) to (_disc_msg_i);

\node[] (_attn_weights_i) at (4.5, -5) {\begin{tabular}{c}Attention \\ Weights $w_{ij}$ \end{tabular}};

\node[circle, draw=black] (_mult_attn) at (2, -5.5) {$\times$};
\node[] () at (0.5, -5.5) {\begin{tabular}{c}Multiplicative \\ Attention $W_a$ \\ with Softmax\end{tabular}};

\draw[->, draw=nice-red, rounded corners=5pt, line width=0.3mm] (_attn_logits_i) -| (_mult_attn);
\draw[->, draw=nice-red, line width=0.3mm] (_mult_attn) to (_attn_weights_i);

\node[] (_disc_msg_j) at (7, -6.1) {Disc. Message $m_j$};
\node[] (_attn_weights_j) at (4.5, -6.1) {\begin{tabular}{c}Attention \\ Weights $w_{ji}$ \end{tabular}};

\draw[->, draw=nice-green, line width=0.3mm] (_mult_attn) to (_attn_weights_j);

\node[] () at (0, -11) {Agent $j$};
\node[rectangle, draw=nice-green, minimum height=0.6cm, minimum width=0.6cm,] (_o_j) at (1, -11) {$o_j$};
\node[rectangle, draw=gray, fill=gray!30] (_obs_encoder_j) at (3.5, -11) {\begin{tabular}{c}Observation \\ Encoder $f_o$ \end{tabular}};
\node[rectangle, draw=nice-green, minimum height=0.6cm, minimum width=0.6cm,] (_e_j) at (6, -11) {$e_j$};
\node[rectangle, draw=yellow, fill=yellow!30] (_attn_encoder_j) at (4.5, -9.5) {\begin{tabular}{c}Attention \\ Encoder $f_a$\end{tabular}};
\node[rectangle, draw=nice-green, fill=nice-green!30] (_msg_encoder_j) at (7, -9.5) {\begin{tabular}{c}Message \\ Encoder $f_m$\end{tabular}};

\draw[->, draw=nice-green, line width=0.3mm] (_o_j) to (_obs_encoder_j);
\draw[->, draw=nice-green, line width=0.3mm] (_obs_encoder_j) to (_e_j);
\draw[->, draw=nice-green, line width=0.3mm] (_e_j) to (_attn_encoder_j);
\draw[->, draw=nice-green, line width=0.3mm] (_e_j) to (_msg_encoder_j);

\node[rectangle, draw=nice-green] (_attn_logits_j) at (4.5, -8) {\begin{tabular}{c}Attention \\ Logits $c_j$ \end{tabular}};
\node[rectangle, draw=nice-green] (_msg_logits_j) at (7, -8) {\begin{tabular}{c}Message \\ Logits $\mu_j$ \end{tabular}};

\draw[->, draw=nice-green, rounded corners=5pt, line width=0.3mm] (_attn_logits_j) -| (_mult_attn);

\draw[->, draw=nice-green, line width=0.3mm] (_attn_encoder_j) to (_attn_logits_j);
\draw[->, draw=nice-green, line width=0.3mm] (_msg_encoder_j) to (_msg_logits_j);

\node[circle, draw=nice-blue, fill=nice-blue!70] (_argmax_j) at (7, -6.9) {};
\node[] () at (8, -6.9) {$\arg \max$};

\draw[->, draw=nice-green, line width=0.3mm] (_msg_logits_j) to (_argmax_j);
\draw[->, draw=nice-green, line width=0.3mm] (_argmax_j) to (_disc_msg_j);

\node[rectangle, draw=nice-red] (_aggr_i) at (13, -3) {$m_{aggr,i} = \sum_{j=1,\dots,n} w_{ij} m_j$};
\node[rectangle, draw=nice-green] (_aggr_j) at (13, -8) {$m_{aggr,j} = \sum_{i=1,\dots,n} w_{ji} m_i$};

\draw[->, draw=nice-red, rounded corners=5pt, line width=0.3mm] (_union_i) -| (_aggr_i);
\draw[->, draw=nice-green, rounded corners=5pt, line width=0.3mm] (_union_j) -| (_aggr_j);

\node[circle, draw=black, minimum width=0.5cm] (_concat_i) at (9, -0.5) {};
\node[] () at (9, 0) {concat};
\node[rectangle, draw=nice-red, minimum width=1.5cm] (_aug_i) at (11.5, -0.5) {\begin{tabular}{c}Augmented \\ Embedding $\hat{e}_i$\end{tabular}};
\node[rectangle, draw=nice-blue, fill=nice-blue!30, minimum width=1.5cm] (_policy_i) at (14.5, -0.5) {Action Policy $\pi$};
\node[rectangle, draw=nice-red] (_a_i) at (16.5, -0.5) {$a_i$};

\draw[->, draw=nice-red, rounded corners=5pt, line width=0.3mm] (4.5, -4.7) |- (9.5, -3.7) -- (9.5, -1.5) -- (_concat_i);
\draw[->, draw=nice-red, line width=0.3mm] (_e_i) to (_concat_i);
\draw[->, draw=nice-red, rounded corners=5pt, line width=0.3mm] (_msg_logits_i) -| (_concat_i);
\draw[->, draw=nice-red, line width=0.3mm] (_aggr_i) to (_concat_i);
\draw[->, draw=nice-red, line width=0.3mm] (_concat_i) to (_aug_i);
\draw[->, draw=nice-red, line width=0.3mm] (_aug_i) to (_policy_i);
\draw[->, draw=nice-red, line width=0.3mm] (_policy_i) to (_a_i);

\node[circle, draw=black, minimum width=0.5cm] (_concat_j) at (9, -10.5) {};
\node[] () at (9, -11) {concat};
\node[rectangle, draw=nice-green, minimum width=1.5cm] (_aug_j) at (11.5, -10.5) {\begin{tabular}{c}Augmented \\ Embedding $\hat{e}_j$\end{tabular}};
\node[rectangle, draw=nice-blue, fill=nice-blue!30, minimum width=1.5cm] (_policy_j) at (14.5, -10.5) {Action Policy $\pi$};
\node[rectangle, draw=nice-green] (_a_j) at (16.5, -10.5) {$a_j$};

\draw[->, draw=nice-green, rounded corners=5pt, line width=0.3mm] (4.5, -6.5) |- (9.5, -7.3) -- (9.5, -9.5) -- (_concat_j);
\draw[->, draw=nice-green, line width=0.3mm] (_e_j) to (_concat_j);
\draw[->, draw=nice-green, rounded corners=5pt, line width=0.3mm] (_msg_logits_j) -| (_concat_j);
\draw[->, draw=nice-green, line width=0.3mm] (_aggr_j) to (_concat_j);
\draw[->, draw=nice-green, line width=0.3mm] (_concat_j) to (_aug_j);
\draw[->, draw=nice-green, line width=0.3mm] (_aug_j) to (_policy_j);
\draw[->, draw=nice-green, line width=0.3mm] (_policy_j) to (_a_j);

\end{tikzpicture}

} 
    \caption{Network architecture}
    \label{fig:model}
\end{figure*}
We use a broadcast-and-listen mechanism to achieve agent communication. Instead of building agent-pair specific communication channels, our model has a public `chat room' to allow agents to share information. To selectively receive information, each agent differentiates the importance of messages from other agents by weighting them using attention weights. Then, agents aggregate the publicly broadcast messages using a weighted sum. The aggregated messages are concatenated with other agent-specific vectors to selection actions for agents. \cref{fig:model} shows the network architecture. It demonstrates the information flow between an agent pair, which can be easily vectorized for any number of agents.

In detail, we first pass $n$ observations $\{o_i\}_{i=1}^n$ of $n$ agents through a parameter sharing observation encoder $f_o$ parameterized by $\theta_o$. The observation encoder outputs observation embeddings  $\{e_i\}_{i=1}^n$:
\begin{equation}
    e_i = f_o(o_i; \theta_o), \text{ for } i = 1,\dots,n.
\end{equation}
We then compute the attention logits $\{c_i\}_{i=1}^n$ and the message logits $\{\mu_i\}_{i=1}^n$ with the parameter sharing attention encoder $f_a$ parameterized by $\theta_a$ and the parameter sharing message encoder $f_m$ parameterized by $\theta_m$ for all observation embeddings  $\{e_i\}_{i=1}^n$:
\begin{equation}
    c_i = f_a(e_i; \theta_a), \enspace \mu_i = f_m(e_i; \theta_m) \text{ for } i = 1,\dots,n.
\end{equation}

We use multiplicative attention~\citep{luong2015effective} to compute the attention scores and then softmax to obtain the attention weights. For an arbitrary pair of agents $i$ and $j$, the attention score $s_{ij}$ and the attention weight $w_{ij}$ of agent $i$ towards agent $j$ are:
\begin{equation}
    s_{ij} = c_j^\top W_a c_i \qquad
    w_{ij} = \frac{\exp(s_{ij})}{\sum_{k=1}^n \exp(s_{ik})}.
\end{equation}
The multiplicative attention operation is parameterized by a square matrix $W_a$.
The attention scores and weights can be efficiently computed for all $i, j$ pairs (including $i=j$, i.e. self-attention weights).

Instead of sampling from distributions, we use the $\argmax$ operation to extract discrete messages from message logits for the agents to broadcast. 
We adopt two approaches to formulate discrete messages: one-hot and bit-string. For a given message bandwidth $b$:
\begin{itemize}
    \item \textit{One-hot}: with a message logit vector $\mu_i \in \mathbb{R}^b$, the $k$th entry of $m_{\text{one-hot}, i} \in \{0, 1\}^b$ is given by
    \begin{equation}
        m_{\text{one-hot}, ik} =
        \begin{cases}
            1 & \text{if $k = \argmax_{l = 1,\dots,b} \mu_{il}$,}\\
            0 & \text{otherwise.}
        \end{cases}
    \end{equation}
    The resulting vocabulary size $|\mathcal{V}|$ of one-hot encoded messages is $b$. An example of an one-hot encoded message with bandwidth $b=5$ can be $m_{\text{one-hot}} = [0,0,1,0,0]$.
    \item \textit{Bit-string}: we first need a message logit vector $\mu_i \in \mathbb{R}^{2b}$, the $k$th entry of $m_{\text{bit-string}, i} \in \{0, 1\}^b$ is given by
    \begin{equation}
        m_{\text{bit-string}, ik} =
        \begin{cases}
            1 & \text{if $k = \argmax_{l\in\{k, k+b\}} \mu_{il}$,}\\
            0 & \text{otherwise.}
        \end{cases}
    \end{equation}
    The resulting vocabulary size $|\mathcal{V}|$ of bit-string encoded messages is $2^b$. An example of a bit-string encoded message with bandwidth $b=5$ can be $m_{\text{bit-string}} = [1,0,0,1,1]$.
\end{itemize}

With the messages $m_i$ and attention weights $w_{ij}$, agents can aggregate the messages: $m_{aggr, i} = \sum_{j=1}^n w_{ij} m_j$.

For agent $i$, we then concatenate the observation embedding $e_i$, the attention weights $w_i=\{w_{ij}\}_{j=1}^n$, the message logits $\mu_i$ and the aggregated message $m_{aggr, i}$ to form an augmented embedding $\hat{e}_i = [e_i; w_i; \mu_i; m_{aggr, i}]$.

The concatenation creates several skip connections in the computation graph. They compensate for the gradient cut-off at the $\argmax$ operation and boost the gradients of the network components closer to the input head.
The augmented embedding $\hat{e}_i$ is then passed through a parameter sharing action policy $\pi$ parameterized by $\theta_\pi$ to infer action
\begin{equation}
    a_i = \pi(\hat{e}_i;\theta_\pi).
\end{equation}

Continuous message communication can be achieved by circumventing the $\argmax$ operation in the network architecture and use message logits $\mu_i$ as the message to broadcast.

In summary, our model uses one round of communication, making the representational capacity of the communication protocols especially important.
\section{Experiments}
We perform experiments and analysis for discrete message communication by 
(1) analyzing the effects of bandwidth and vocabulary size and comparing with continuous message communication; 
(2) analyzing the importance of self-attention for discrete message communication; 
(3) analyzing positive listening and signaling; and 
(4) introducing human interaction with agents by using discrete message communication. 

We show the results obtained from three environments: Pulling Levers~\citep{sukhbaatar2016learning}, Predator-Prey~\citep{Bohmer2019-zv, li2020deep}, and Multi-Walker~\citep{gupta2017cooperative, terry2020pettingzoo}. These environments have challenging tasks that must be performed cooperatively to achieve high returns. Using only local information cannot achieve high returns. 

We use PPO~\citep{schulman2017proximal} for policy optimization (see Section~\ref{sec:ppo}). A multi-layer perceptron (MLP) baseline (value function) with global state is used for reducing the variance of advantage estimation with GAE~\citep{schulman2015high} during training.

\subsection{Analysis for Bandwidth and Vocabulary Size}
We compare the performance of communication protocols with various bandwidth and vocabulary size. The metric used is the average return. Average return indirectly measures the effects of communication on behaviors (i.e. positive listening). Continuous message communication is used as a baseline. 

\subsubsection{Pulling Levers}\label{sec:levers_results}

\begin{figure}[t]
\begin{tikzpicture}

\definecolor{color0}{rgb}{0.86,0.3712,0.34}
\definecolor{color1}{rgb}{0.86,0.554729411764706,0.34}
\definecolor{color2}{rgb}{0.86,0.738258823529412,0.34}
\definecolor{color3}{rgb}{0.798211764705882,0.86,0.34}
\definecolor{color4}{rgb}{0.614682352941176,0.86,0.34}
\definecolor{color5}{rgb}{0.43115294117647,0.86,0.34}
\definecolor{color6}{rgb}{0.34,0.86,0.432376470588235}
\definecolor{color7}{rgb}{0.34,0.86,0.615905882352941}
\definecolor{color8}{rgb}{0.34,0.86,0.799435294117647}
\definecolor{color9}{rgb}{0.34,0.737035294117647,0.86}
\definecolor{color10}{rgb}{0.34,0.553505882352941,0.86}
\definecolor{color11}{rgb}{0.34,0.369976470588235,0.86}
\definecolor{color12}{rgb}{0.49355294117647,0.34,0.86}
\definecolor{color13}{rgb}{0.677082352941176,0.34,0.86}
\definecolor{color14}{rgb}{0.86,0.34,0.859388235294117}
\definecolor{color15}{rgb}{0.86,0.34,0.675858823529411}
\definecolor{color16}{rgb}{0.86,0.34,0.492329411764706}

\node[scale=0.65] () at (0, -0.47) {$|\mathcal{V}|$};

\begin{axis}[
legend cell align={left},
legend style={fill opacity=0.8, draw opacity=1, text opacity=1, draw=white!80!black},
tick align=outside,
tick pos=left,
x grid style={white!69.0196078431373!black},
xlabel={Communication Protocols},
xmin=-1, xmax=17,
xtick style={color=black},
xtick={0,1,2,3,4,5,6,7,8,9,10,11,12,13,14,15,16},
xticklabels={no\\0,c8\\$\infty$,c16\\$\infty$,c32\\$\infty$,c64\\$\infty$,b4\\$16$,b6\\$64$,b8\\$256$,b10\\$2^{10}$,b16\\$2^{16}$,b32\\$2^{32}$,o8\\$8$,o16\\$16$,o32\\$32$,o64\\$64$,o128\\$128$,o256\\$256$},
xticklabel style={align=center},
y grid style={white!69.0196078431373!black},
ylabel={Average Return},
ymin=0.66, ymax=0.9,
ytick style={color=black},
ytick={0.65,0.7,0.75,0.8,0.85,0.9},
yticklabels={0.65,0.70,0.75,0.80,0.85,0.90},
every tick label/.append style={scale=0.53},
every axis label/.append style={scale=0.8},
legend style={nodes={scale=0.7, transform shape}},
width=3.38in,
height=2.2in,
]
\draw[draw=none,fill=color0] (axis cs:-0.4,0) rectangle (axis cs:0.4,0.698755848736496);
\draw[draw=none,fill=color1] (axis cs:0.6,0) rectangle (axis cs:1.4,0.867422073999779);
\draw[draw=none,fill=color2] (axis cs:1.6,0) rectangle (axis cs:2.4,0.870324899591157);
\draw[draw=none,fill=color3] (axis cs:2.6,0) rectangle (axis cs:3.4,0.873476002957951);
\draw[draw=none,fill=color4] (axis cs:3.6,0) rectangle (axis cs:4.4,0.861949376603286);
\draw[draw=none,fill=color5] (axis cs:4.6,0) rectangle (axis cs:5.4,0.756088112946136);
\draw[draw=none,fill=color6] (axis cs:5.6,0) rectangle (axis cs:6.4,0.770543753164072);
\draw[draw=none,fill=color7] (axis cs:6.6,0) rectangle (axis cs:7.4,0.790537079482189);
\draw[draw=none,fill=color8] (axis cs:7.6,0) rectangle (axis cs:8.4,0.793018741782761);
\draw[draw=none,fill=color9] (axis cs:8.6,0) rectangle (axis cs:9.4,0.806977998079031);
\draw[draw=none,fill=color10] (axis cs:9.6,0) rectangle (axis cs:10.4,0.833606473765183);
\draw[draw=none,fill=color11] (axis cs:10.6,0) rectangle (axis cs:11.4,0.763314342017357);
\draw[draw=none,fill=color12] (axis cs:11.6,0) rectangle (axis cs:12.4,0.771555999966);
\draw[draw=none,fill=color13] (axis cs:12.6,0) rectangle (axis cs:13.4,0.789217968159525);
\draw[draw=none,fill=color14] (axis cs:13.6,0) rectangle (axis cs:14.4,0.788212318064752);
\draw[draw=none,fill=color15] (axis cs:14.6,0) rectangle (axis cs:15.4,0.774491445163155);
\draw[draw=none,fill=color16] (axis cs:15.6,0) rectangle (axis cs:16.4,0.768117627179151);
\path [draw=black, semithick]
(axis cs:0,0.697091916082584)
--(axis cs:0,0.700419781390407);

\path [draw=black, semithick]
(axis cs:1,0.852838474798766)
--(axis cs:1,0.882005673200792);

\path [draw=black, semithick]
(axis cs:2,0.858369234805226)
--(axis cs:2,0.882280564377087);

\path [draw=black, semithick]
(axis cs:3,0.859103361245739)
--(axis cs:3,0.887848644670163);

\path [draw=black, semithick]
(axis cs:4,0.851144397589439)
--(axis cs:4,0.872754355617132);

\path [draw=black, semithick]
(axis cs:5,0.744786213472278)
--(axis cs:5,0.767390012419995);

\path [draw=black, semithick]
(axis cs:6,0.761086847391817)
--(axis cs:6,0.780000658936327);

\path [draw=black, semithick]
(axis cs:7,0.78084750787852)
--(axis cs:7,0.800226651085857);

\path [draw=black, semithick]
(axis cs:8,0.779440927423098)
--(axis cs:8,0.806596556142424);

\path [draw=black, semithick]
(axis cs:9,0.796956200385894)
--(axis cs:9,0.816999795772169);

\path [draw=black, semithick]
(axis cs:10,0.817617627811541)
--(axis cs:10,0.849595319718824);

\path [draw=black, semithick]
(axis cs:11,0.756307551255004)
--(axis cs:11,0.770321132779709);

\path [draw=black, semithick]
(axis cs:12,0.764223322731175)
--(axis cs:12,0.778888677200826);

\path [draw=black, semithick]
(axis cs:13,0.778244503864886)
--(axis cs:13,0.800191432454164);

\path [draw=black, semithick]
(axis cs:14,0.776502378974747)
--(axis cs:14,0.799922257154756);

\path [draw=black, semithick]
(axis cs:15,0.760648614152267)
--(axis cs:15,0.788334276174043);

\path [draw=black, semithick]
(axis cs:16,0.757655684498805)
--(axis cs:16,0.778579569859497);

\addplot [semithick, black, mark=-, mark size=3, mark options={solid}, only marks, forget plot]
table {%
0 0.697091916082584
1 0.852838474798766
2 0.858369234805226
3 0.859103361245739
4 0.851144397589439
5 0.744786213472278
6 0.761086847391817
7 0.78084750787852
8 0.779440927423098
9 0.796956200385894
10 0.817617627811541
11 0.756307551255004
12 0.764223322731175
13 0.778244503864886
14 0.776502378974747
15 0.760648614152267
16 0.757655684498805
};
\addplot [semithick, black, mark=-, mark size=3, mark options={solid}, only marks, forget plot]
table {%
0 0.700419781390407
1 0.882005673200792
2 0.882280564377087
3 0.887848644670163
4 0.872754355617132
5 0.767390012419995
6 0.780000658936327
7 0.800226651085857
8 0.806596556142424
9 0.816999795772169
10 0.849595319718824
11 0.770321132779709
12 0.778888677200826
13 0.800191432454164
14 0.799922257154756
15 0.788334276174043
16 0.778579569859497
};
\addplot [semithick, red, dashed]
table {%
-1 0.67232
18 0.67232
};
\addlegendentry{Random Action Expected Return}
\end{axis}

\end{tikzpicture}
    \caption{Returns of Pulling Levers (5 levers, 20 total participants) with various communication protocols.}
    \label{fig:levers_5_20}
\end{figure}
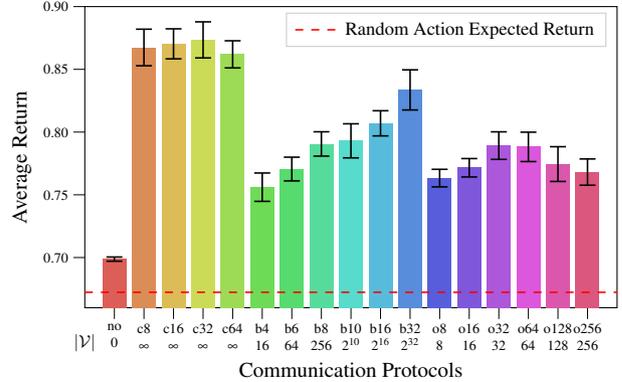

\begin{figure}[t]
\begin{tikzpicture}

\definecolor{color0}{rgb}{0.86,0.3712,0.34}
\definecolor{color1}{rgb}{0.86,0.5662,0.34}
\definecolor{color2}{rgb}{0.86,0.7612,0.34}
\definecolor{color3}{rgb}{0.7638,0.86,0.34}
\definecolor{color4}{rgb}{0.5688,0.86,0.34}
\definecolor{color5}{rgb}{0.3738,0.86,0.34}
\definecolor{color6}{rgb}{0.34,0.86,0.5012}
\definecolor{color7}{rgb}{0.34,0.86,0.6962}
\definecolor{color8}{rgb}{0.34,0.8288,0.86}
\definecolor{color9}{rgb}{0.34,0.6338,0.86}
\definecolor{color10}{rgb}{0.34,0.4388,0.86}
\definecolor{color11}{rgb}{0.4362,0.34,0.86}
\definecolor{color12}{rgb}{0.6312,0.34,0.86}
\definecolor{color13}{rgb}{0.8262,0.34,0.86}
\definecolor{color14}{rgb}{0.86,0.34,0.6988}
\definecolor{color15}{rgb}{0.86,0.34,0.5038}

\node[scale=0.65] () at (0, -0.47) {$|\mathcal{V}|$};

\begin{axis}[
legend cell align={left},
legend style={fill opacity=0.8, draw opacity=1, text opacity=1, draw=white!80!black},
tick align=outside,
tick pos=left,
x grid style={white!69.0196078431373!black},
xlabel={Communication Protocols},
xmin=-1, xmax=16,
xtick style={color=black},
xtick={0,1,2,3,4,5,6,7,8,9,10,11,12,13,14,15},
xticklabels={no\\0,c8\\$\infty$,c16\\$\infty$,c32\\$\infty$,c64\\$\infty$,b4\\$16$,b6\\$64$,b8\\$256$,b10\\$2^{10}$,b16\\$2^{16}$,b32\\$2^{32}$,o8\\$8$,o16\\$16$,o32\\$32$,o64\\$64$,o128\\$128$},
xticklabel style={align=center},
y grid style={white!69.0196078431373!black},
ylabel={Average Return},
ymin=0.66, ymax=0.9,
ytick style={color=black},
ytick={0.65,0.7,0.75,0.8,0.85,0.9},
yticklabels={0.65,0.70,0.75,0.80,0.85,0.90},
every tick label/.append style={scale=0.53},
every axis label/.append style={scale=0.8},
legend style={nodes={scale=0.7, transform shape}},
width=3.38in,
height=2.2in,
]
\draw[draw=none,fill=color0] (axis cs:-0.4,0) rectangle (axis cs:0.4,0.664555151543999);
\draw[draw=none,fill=color1] (axis cs:0.6,0) rectangle (axis cs:1.4,0.775222870198641);
\draw[draw=none,fill=color2] (axis cs:1.6,0) rectangle (axis cs:2.4,0.786151867198191);
\draw[draw=none,fill=color3] (axis cs:2.6,0) rectangle (axis cs:3.4,0.791408991831634);
\draw[draw=none,fill=color4] (axis cs:3.6,0) rectangle (axis cs:4.4,0.78234766425554);
\draw[draw=none,fill=color5] (axis cs:4.6,0) rectangle (axis cs:5.4,0.708924049537182);
\draw[draw=none,fill=color6] (axis cs:5.6,0) rectangle (axis cs:6.4,0.718445771948763);
\draw[draw=none,fill=color7] (axis cs:6.6,0) rectangle (axis cs:7.4,0.728271048967692);
\draw[draw=none,fill=color8] (axis cs:7.6,0) rectangle (axis cs:8.4,0.753466185007947);
\draw[draw=none,fill=color9] (axis cs:8.6,0) rectangle (axis cs:9.4,0.76933878421406);
\draw[draw=none,fill=color10] (axis cs:9.6,0) rectangle (axis cs:10.4,0.707219630460097);
\draw[draw=none,fill=color11] (axis cs:10.6,0) rectangle (axis cs:11.4,0.714393420156567);
\draw[draw=none,fill=color12] (axis cs:11.6,0) rectangle (axis cs:12.4,0.723173906926536);
\draw[draw=none,fill=color13] (axis cs:12.6,0) rectangle (axis cs:13.4,0.726413077816216);
\draw[draw=none,fill=color14] (axis cs:13.6,0) rectangle (axis cs:14.4,0.736898056830062);
\draw[draw=none,fill=color15] (axis cs:14.6,0) rectangle (axis cs:15.4,0.71140566723049);
\path [draw=black, semithick]
(axis cs:0,0.663412438232369)
--(axis cs:0,0.66569786485563);

\path [draw=black, semithick]
(axis cs:1,0.769927173772832)
--(axis cs:1,0.78051856662445);

\path [draw=black, semithick]
(axis cs:2,0.781707201591174)
--(axis cs:2,0.790596532805208);

\path [draw=black, semithick]
(axis cs:3,0.783977810388528)
--(axis cs:3,0.798840173274741);

\path [draw=black, semithick]
(axis cs:4,0.777238120742208)
--(axis cs:4,0.787457207768872);

\path [draw=black, semithick]
(axis cs:5,0.704112931067837)
--(axis cs:5,0.713735168006528);

\path [draw=black, semithick]
(axis cs:6,0.709118821008253)
--(axis cs:6,0.727772722889272);

\path [draw=black, semithick]
(axis cs:7,0.724200624760091)
--(axis cs:7,0.732341473175292);

\path [draw=black, semithick]
(axis cs:8,0.745611308920602)
--(axis cs:8,0.761321061095292);

\path [draw=black, semithick]
(axis cs:9,0.762138636030906)
--(axis cs:9,0.776538932397215);

\path [draw=black, semithick]
(axis cs:10,0.699899400175097)
--(axis cs:10,0.714539860745097);

\path [draw=black, semithick]
(axis cs:11,0.704358591593639)
--(axis cs:11,0.724428248719496);

\path [draw=black, semithick]
(axis cs:12,0.716468126851907)
--(axis cs:12,0.729879687001165);

\path [draw=black, semithick]
(axis cs:13,0.709054670105143)
--(axis cs:13,0.743771485527289);

\path [draw=black, semithick]
(axis cs:14,0.719544933230202)
--(axis cs:14,0.754251180429923);

\path [draw=black, semithick]
(axis cs:15,0.702835576169793)
--(axis cs:15,0.719975758291188);

\addplot [semithick, black, mark=-, mark size=3, mark options={solid}, only marks, forget plot]
table {%
0 0.663412438232369
1 0.769927173772832
2 0.781707201591174
3 0.783977810388528
4 0.777238120742208
5 0.704112931067837
6 0.709118821008253
7 0.724200624760091
8 0.745611308920602
9 0.762138636030906
10 0.699899400175097
11 0.704358591593639
12 0.716468126851907
13 0.709054670105143
14 0.719544933230202
15 0.702835576169793
};
\addplot [semithick, black, mark=-, mark size=3, mark options={solid}, only marks, forget plot]
table {%
0 0.66569786485563
1 0.78051856662445
2 0.790596532805208
3 0.798840173274741
4 0.787457207768872
5 0.713735168006528
6 0.727772722889272
7 0.732341473175292
8 0.761321061095292
9 0.776538932397215
10 0.714539860745097
11 0.724428248719496
12 0.729879687001165
13 0.743771485527289
14 0.754251180429923
15 0.719975758291188
};
\addplot [semithick, red, dashed]
table {%
-1 0.67232
17 0.67232
};
\addlegendentry{Random Action Expected Return}
\end{axis}

\end{tikzpicture}
    \caption{Returns of Pulling Levers (5 levers, 100 total participants) with various communication protocols.}
    \label{fig:levers_5_100}
\end{figure}
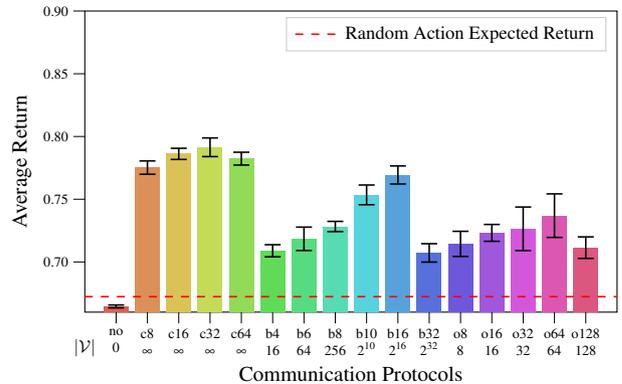

Pulling Levers~\citep{sukhbaatar2016learning} is a simple multi-agent multi-armed bandit task that requires communication to achieve high a score. There are $n$ levers in the task. In each round, $n$ agents are randomly sampled from a total of $N$ participants. The action of each agent is to select a lever to pull. The reward of one round is given by the ratio between the number of unique levers pulled by the agents and the number of levers $n$. The maximum reward per round is therefore $1$, meaning each pulled lever is unique in that step. The theoretical expected reward of randomly pulling levers is $1 - (\frac{n-1}{n})^n$. The observation $o_i$ for each agent is the one-hot encoding of its ID number. Choosing levers without communication in this game will be similar to acting randomly.

A \textit{zero baseline} (value function) is used for training the pulling levers environment, since the global state (observation) is not related to return. Each episode consists of 50 rounds, and reward for one episode is summed over rewards from each round. Results are averaged over $5$ random seeds.

\cref{fig:levers_5_20} shows the normalized average returns of various communication protocols with $n=5$ and $N=20$. The communication protocols and their vocabulary sizes $|\mathcal{V}|$ are labeled on the horizontal axis: `\texttt{no}' stands for no communication, `\texttt{c}' means continuous message, `\texttt{b}' means bit-string, `\texttt{o}' means one-hot, and the number following the letter is the bandwidth $b$ of communication. A red dashed line is over-plotted to indicate the expected return with random action. 

From \cref{fig:levers_5_20}, No communication shows the performance close to acting randomly. Continuous message works the best. The performance differences between continuous message bandwidths are not significant, since continuous message with any bandwidth can effectively encode an infinite number of meanings. Discrete message communication shows inferior performance due to its limited representational capacity. Generally, the performance of discrete message communication increases when the vocabulary size increases. This is particularly true for bit-string encoding. The 32-bit-string works the best among discrete communication, since its vocabulary size is $2^{32} \approx 4 \times 10^9$. This size is beyond human’s cognitive range. The average vocabulary size of native English speakers is around 20000, while 6000 to 7000 are sufficient for understanding most communication~\citep{rosenberg2008human}. The vocabulary sizes of the one-hot messages are within a human’s cognitive range, and their performance are not too far off comparing with continuous and bit-string messaging. The performance of one-hot messaging drops as the bandwidth increases beyond 128. This result indicates that overly large bandwidths are harmful for learning emergent communication protocols. 

\cref{fig:levers_5_100} shows the normalized average returns for $n=5$ and $N=100$. The trend of the results is consistent with $N=20$.

\subsubsection{Predatory-Prey}\label{sec:predprey_results}
\begin{figure}
    \centering
    \includegraphics[width=0.18\textwidth]{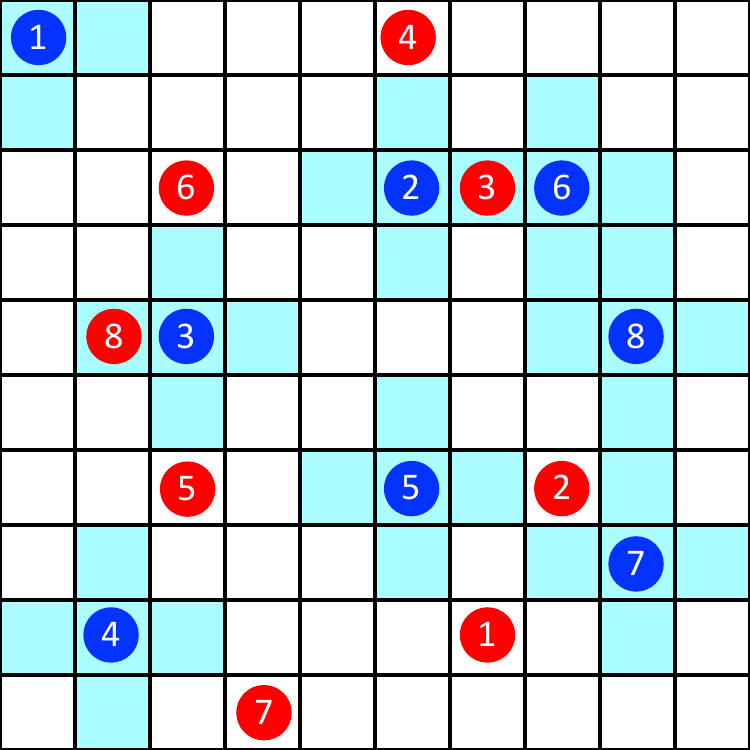}
    \caption{Predators are marked in blue, and prey are marked in red. The cyan grids are the capture range of predators. An example of successful capture is predator 2 and 6 capturing prey 3. An example of a single-agent capture attempt that results in a penalty is predator 3 capturing prey 8 alone.}
    \label{fig:predprey_env}
\end{figure}
We use a Predatory-Prey environment similar to that described by \citet{Bohmer2019-zv} and \citet{li2020deep}. The environment consists of a $7\times7$ grid world with 4 predators and 4 prey. We control the movement of predators to capture prey. The prey move by hard-coded and randomized rules to avoid predators. If a prey is captured, the agents receives a reward of $10$. However, the environment penalizes any single-agent attempt to capture prey with a negative reward $-0.5$; at least two agents are required to be present in the neighboring grid cells of a prey for a successful capture. \cref{fig:predprey_env} illustrates the environment and the reward mechanism. The agents have a vision range of 2 grids from itself. We engineer the agent's observation so that we can remove other agents positions away from an agent’s field of view. That means an agent can now only see the prey but not the other agents. The combination of agent invisibility and the single-agent capture-attempt penalty makes the task even harder. Cooperation is necessary to achieve a high return in this environment. We set the episode length to $50$ steps, and impose a step cost of $-0.1$.

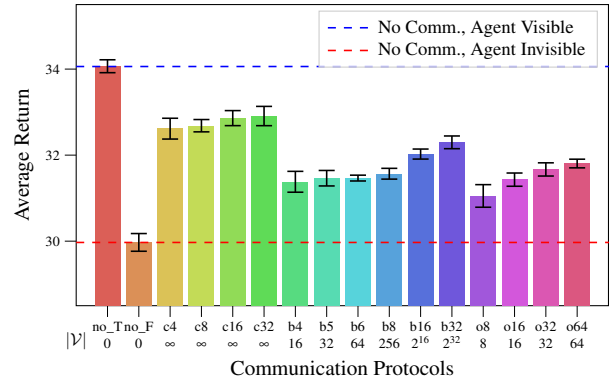
\begin{figure}
\begin{tikzpicture}

\definecolor{color0}{rgb}{0.86,0.3712,0.34}
\definecolor{color1}{rgb}{0.86,0.5662,0.34}
\definecolor{color2}{rgb}{0.86,0.7612,0.34}
\definecolor{color3}{rgb}{0.7638,0.86,0.34}
\definecolor{color4}{rgb}{0.5688,0.86,0.34}
\definecolor{color5}{rgb}{0.3738,0.86,0.34}
\definecolor{color6}{rgb}{0.34,0.86,0.5012}
\definecolor{color7}{rgb}{0.34,0.86,0.6962}
\definecolor{color8}{rgb}{0.34,0.8288,0.86}
\definecolor{color9}{rgb}{0.34,0.6338,0.86}
\definecolor{color10}{rgb}{0.34,0.4388,0.86}
\definecolor{color11}{rgb}{0.4362,0.34,0.86}
\definecolor{color12}{rgb}{0.6312,0.34,0.86}
\definecolor{color13}{rgb}{0.8262,0.34,0.86}
\definecolor{color14}{rgb}{0.86,0.34,0.6988}
\definecolor{color15}{rgb}{0.86,0.34,0.5038}

\node[scale=0.65] () at (0, -0.47) {$|\mathcal{V}|$};

\begin{axis}[
legend cell align={left},
legend style={fill opacity=0.7, draw opacity=1, text opacity=1, draw=white!80!black},
tick align=outside,
tick pos=left,
x grid style={white!69.0196078431373!black},
xlabel={Communication Protocols},
xmin=-1, xmax=16,
xtick style={color=black},
xtick={0,1,2,3,4,5,6,7,8,9,10,11,12,13,14,15},
xticklabels={no\_T\\0,no\_F\\0,c4\\$\infty$,c8\\$\infty$,c16\\$\infty$,c32\\$\infty$,b4\\$16$,b5\\$32$,b6\\$64$,b8\\$256$,b16\\$2^{16}$,b32\\$2^{32}$,o8\\$8$,o16\\$16$,o32\\$32$,o64\\$64$},
xticklabel style={align=center},
y grid style={white!69.0196078431373!black},
ylabel={Average Return},
ymin=28.5, ymax=35.5,
ytick style={color=black},
every tick label/.append style={scale=0.53},
every axis label/.append style={scale=0.8},
legend style={nodes={scale=0.7, transform shape}},
width=3.38in,
height=2.2in,
]
\draw[draw=none,fill=color0] (axis cs:-0.4,0) rectangle (axis cs:0.4,34.0665469625711);
\draw[draw=none,fill=color1] (axis cs:0.6,0) rectangle (axis cs:1.4,29.9712101970206);
\draw[draw=none,fill=color2] (axis cs:1.6,0) rectangle (axis cs:2.4,32.6154410837646);
\draw[draw=none,fill=color3] (axis cs:2.6,0) rectangle (axis cs:3.4,32.6842143898773);
\draw[draw=none,fill=color4] (axis cs:3.6,0) rectangle (axis cs:4.4,32.8602479459451);
\draw[draw=none,fill=color5] (axis cs:4.6,0) rectangle (axis cs:5.4,32.9087645384683);
\draw[draw=none,fill=color6] (axis cs:5.6,0) rectangle (axis cs:6.4,31.3804071635213);
\draw[draw=none,fill=color7] (axis cs:6.6,0) rectangle (axis cs:7.4,31.4636534355067);
\draw[draw=none,fill=color8] (axis cs:7.6,0) rectangle (axis cs:8.4,31.4665703988891);
\draw[draw=none,fill=color9] (axis cs:8.6,0) rectangle (axis cs:9.4,31.5676292612859);
\draw[draw=none,fill=color10] (axis cs:9.6,0) rectangle (axis cs:10.4,32.024841266935);
\draw[draw=none,fill=color11] (axis cs:10.6,0) rectangle (axis cs:11.4,32.2974610540544);
\draw[draw=none,fill=color12] (axis cs:11.6,0) rectangle (axis cs:12.4,31.0514136281669);
\draw[draw=none,fill=color13] (axis cs:12.6,0) rectangle (axis cs:13.4,31.4302975824806);
\draw[draw=none,fill=color14] (axis cs:13.6,0) rectangle (axis cs:14.4,31.6675999017907);
\draw[draw=none,fill=color15] (axis cs:14.6,0) rectangle (axis cs:15.4,31.804926554153);
\path [draw=black, semithick]
(axis cs:0,33.9164919924203)
--(axis cs:0,34.2166019327218);

\path [draw=black, semithick]
(axis cs:1,29.7654508958223)
--(axis cs:1,30.1769694982188);

\path [draw=black, semithick]
(axis cs:2,32.3735866966612)
--(axis cs:2,32.8572954708681);

\path [draw=black, semithick]
(axis cs:3,32.5411002019049)
--(axis cs:3,32.8273285778497);

\path [draw=black, semithick]
(axis cs:4,32.6853762757539)
--(axis cs:4,33.0351196161363);

\path [draw=black, semithick]
(axis cs:5,32.684540321747)
--(axis cs:5,33.1329887551896);

\path [draw=black, semithick]
(axis cs:6,31.1379143482224)
--(axis cs:6,31.6228999788202);

\path [draw=black, semithick]
(axis cs:7,31.2836309578772)
--(axis cs:7,31.6436759131362);

\path [draw=black, semithick]
(axis cs:8,31.3988955603834)
--(axis cs:8,31.5342452373948);

\path [draw=black, semithick]
(axis cs:9,31.4422671475875)
--(axis cs:9,31.6929913749842);

\path [draw=black, semithick]
(axis cs:10,31.9083466339571)
--(axis cs:10,32.1413358999129);

\path [draw=black, semithick]
(axis cs:11,32.1495452742194)
--(axis cs:11,32.4453768338894);

\path [draw=black, semithick]
(axis cs:12,30.7888623888572)
--(axis cs:12,31.3139648674767);

\path [draw=black, semithick]
(axis cs:13,31.2765391062531)
--(axis cs:13,31.5840560587082);

\path [draw=black, semithick]
(axis cs:14,31.5140022572629)
--(axis cs:14,31.8211975463185);

\path [draw=black, semithick]
(axis cs:15,31.7041678389236)
--(axis cs:15,31.9056852693824);

\addplot [semithick, black, mark=-, mark size=3, mark options={solid}, only marks, forget plot]
table {%
0 33.9164919924203
1 29.7654508958223
2 32.3735866966612
3 32.5411002019049
4 32.6853762757539
5 32.684540321747
6 31.1379143482224
7 31.2836309578772
8 31.3988955603834
9 31.4422671475875
10 31.9083466339571
11 32.1495452742194
12 30.7888623888572
13 31.2765391062531
14 31.5140022572629
15 31.7041678389236
};
\addplot [semithick, black, mark=-, mark size=3, mark options={solid}, only marks, forget plot]
table {%
0 34.2166019327218
1 30.1769694982188
2 32.8572954708681
3 32.8273285778497
4 33.0351196161363
5 33.1329887551896
6 31.6228999788202
7 31.6436759131362
8 31.5342452373948
9 31.6929913749842
10 32.1413358999129
11 32.4453768338894
12 31.3139648674767
13 31.5840560587082
14 31.8211975463185
15 31.9056852693824
};
\addplot [semithick, blue, dashed]
table {%
-1 34.06
17 34.06
};
\addlegendentry{No Comm., Agent Visible}
\addplot [semithick, red, dashed]
table {%
-1 29.97
17 29.97
};
\addlegendentry{No Comm., Agent Invisible}
\end{axis}

\end{tikzpicture}
    \caption{Returns of the 4-agent Predator-Prey with various communication protocols.}
    \label{fig:predprey_results}
\end{figure}

The results are shown in \cref{fig:predprey_results}. Results are averaged over 5 random seeds. We ran experiments for various communication protocols. In particular, no communication \textbf{with} agent visibility (\texttt{no\_T}) and no communication \textbf{without} agent visibility (\texttt{no\_F}) form the performance upper-bound and lower-bound respectively. In the former setting, each agent has the richest information about the other agents, whereas in the latter setting, each agent has the scarcest information. Agents are set to invisible for all the other communication protocols. The dashed lines mark the upper and lower bounds. (The labels on the horizontal axis follow the convention defined in \cref{sec:levers_results} and \cref{fig:levers_5_20}.)

The performance of all the communicative methods falls within the upper and lower bounds. We can see similar increasing patterns for increasing bandwidth as in \cref{sec:levers_results} for Pulling Levers, which indicates that richer information transmission comes with higher bandwidth (up to a certain limit). Continuous messaging performs the best, followed by bit-string messaging. Bit-string messaging matches continuous messaging performance as its vocabulary size increases to $2^{32}$. One-hot messaging is slightly better than bit-string messaging when their vocabulary sizes are same. All the discrete communication protocols outperform no communication by a large margin.

\subsubsection{Multi-Walker}
\begin{figure}
    \centering
    \includegraphics[width=0.25\textwidth]{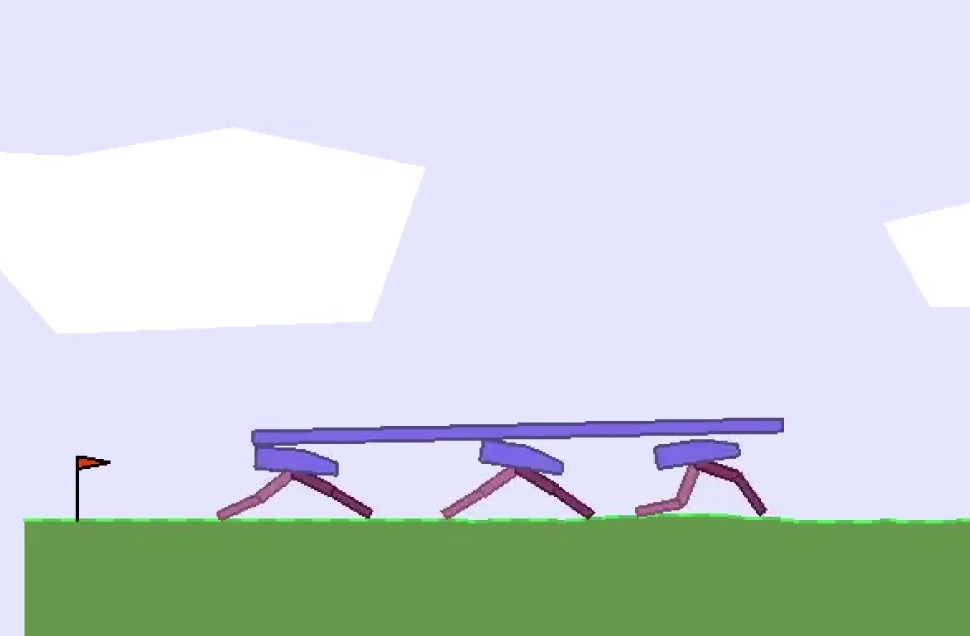}
    \caption{An illustration of the Multi-Walker environment with 3 agents.}
    \label{fig:multi-walker_env}
\end{figure}

The previous two domains consist of discrete observation and action spaces. We would like to examine the consistency of discrete message communication in a more challenging continuous observation and action space task. 
In the Multi-Walker environment~\citep{gupta2017cooperative, terry2020pettingzoo}, $n$ bipedal robot agents try to carry a bar-shaped package and move forward as far as possible as illustrated in \cref{fig:multi-walker_env}. The agents receive positive rewards for moving forward and negative reward for moving backward. Large negative rewards are given if the an agent falls or the package falls. The task is a combination of learning robotic locomotion and inter-agent cooperation.

In the original observation design, each agent receives an observation composed of physical properties of its legs and joints, as well as LIDAR readings from the space immediately in front and below the robot. The original observation also includes information about neighboring walkers, and the package. To emphasize the necessity of communication, we engineer the observation space to remove LIDAR readings and information about neighboring walkers. \textit{The resulting engineered observation therefore is only composed of the agent's own physical properties and information about the package.} The information about the other agents is now only accessible from communication.

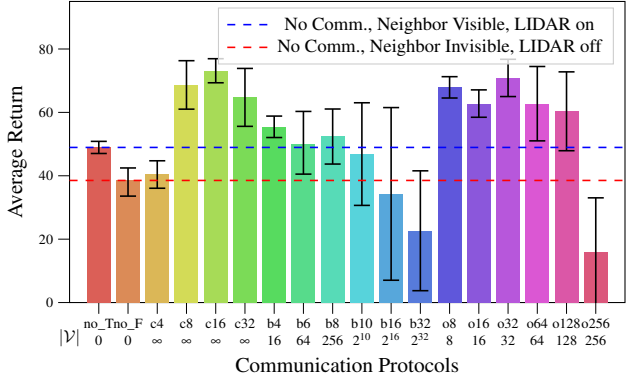
\begin{figure}[t]
\begin{tikzpicture}

\definecolor{color0}{rgb}{0.86,0.3712,0.34}
\definecolor{color1}{rgb}{0.86,0.544533333333333,0.34}
\definecolor{color2}{rgb}{0.86,0.717866666666667,0.34}
\definecolor{color3}{rgb}{0.8288,0.86,0.34}
\definecolor{color4}{rgb}{0.655466666666666,0.86,0.34}
\definecolor{color5}{rgb}{0.482133333333333,0.86,0.34}
\definecolor{color6}{rgb}{0.34,0.86,0.3712}
\definecolor{color7}{rgb}{0.34,0.86,0.544533333333333}
\definecolor{color8}{rgb}{0.34,0.86,0.717866666666667}
\definecolor{color9}{rgb}{0.34,0.8288,0.86}
\definecolor{color10}{rgb}{0.34,0.655466666666666,0.86}
\definecolor{color11}{rgb}{0.34,0.482133333333333,0.86}
\definecolor{color12}{rgb}{0.3712,0.34,0.86}
\definecolor{color13}{rgb}{0.544533333333333,0.34,0.86}
\definecolor{color14}{rgb}{0.717866666666666,0.34,0.86}
\definecolor{color15}{rgb}{0.86,0.34,0.8288}
\definecolor{color16}{rgb}{0.86,0.34,0.655466666666666}
\definecolor{color17}{rgb}{0.86,0.34,0.482133333333333}

\node[scale=0.65] () at (0, -0.47) {$|\mathcal{V}|$};

\begin{axis}[
legend style={fill opacity=0.8, draw opacity=1, text opacity=1, draw=white!80!black},
tick align=outside,
tick pos=left,
x grid style={white!69.0196078431373!black},
xlabel={Communication Protocols},
xmin=-1, xmax=18,
xtick style={color=black},
xtick={0,1,2,3,4,5,6,7,8,9,10,11,12,13,14,15,16,17},
xticklabels={no\_T\\0,no\_F\\0,c4\\$\infty$,c8\\$\infty$,c16\\$\infty$,c32\\$\infty$,b4\\$16$,b6\\$64$,b8\\$256$,b10\\$2^{10}$,b16\\$2^{16}$,b32\\$2^{32}$,o8\\$8$,o16\\$16$,o32\\$32$,o64\\$64$,o128\\$128$,o256\\$256$},
xticklabel style={align=center},
y grid style={white!69.0196078431373!black},
ylabel={Average Return},
ymin=0, ymax=95,
ytick style={color=black},
every tick label/.append style={scale=0.53},
every axis label/.append style={scale=0.8},
legend style={nodes={scale=0.7, transform shape}},
width=3.5in,
height=2.2in,
]
\draw[draw=none,fill=color0] (axis cs:-0.4,0) rectangle (axis cs:0.4,48.9440278918437);
\draw[draw=none,fill=color1] (axis cs:0.6,0) rectangle (axis cs:1.4,38.5251531348666);
\draw[draw=none,fill=color2] (axis cs:1.6,0) rectangle (axis cs:2.4,40.3994210122897);
\draw[draw=none,fill=color3] (axis cs:2.6,0) rectangle (axis cs:3.4,68.6632155724459);
\draw[draw=none,fill=color4] (axis cs:3.6,0) rectangle (axis cs:4.4,73.1397325781034);
\draw[draw=none,fill=color5] (axis cs:4.6,0) rectangle (axis cs:5.4,64.7314203308059);
\draw[draw=none,fill=color6] (axis cs:5.6,0) rectangle (axis cs:6.4,55.45075009258);
\draw[draw=none,fill=color7] (axis cs:6.6,0) rectangle (axis cs:7.4,49.9130804905345);
\draw[draw=none,fill=color8] (axis cs:7.6,0) rectangle (axis cs:8.4,52.3820098980336);
\draw[draw=none,fill=color9] (axis cs:8.6,0) rectangle (axis cs:9.4,46.8447223779467);
\draw[draw=none,fill=color10] (axis cs:9.6,0) rectangle (axis cs:10.4,34.2788118346906);
\draw[draw=none,fill=color11] (axis cs:10.6,0) rectangle (axis cs:11.4,22.6515352637727);
\draw[draw=none,fill=color12] (axis cs:11.6,0) rectangle (axis cs:12.4,67.9015715625081);
\draw[draw=none,fill=color13] (axis cs:12.6,0) rectangle (axis cs:13.4,62.7822084032171);
\draw[draw=none,fill=color14] (axis cs:13.6,0) rectangle (axis cs:14.4,70.8746051053793);
\draw[draw=none,fill=color15] (axis cs:14.6,0) rectangle (axis cs:15.4,62.7482438871572);
\draw[draw=none,fill=color16] (axis cs:15.6,0) rectangle (axis cs:16.4,60.3462835411104);
\draw[draw=none,fill=color17] (axis cs:16.6,0) rectangle (axis cs:17.4,15.8426668921398);
\path [draw=black, semithick]
(axis cs:0,47.0319880887532)
--(axis cs:0,50.8560676949342);

\path [draw=black, semithick]
(axis cs:1,33.5834137395823)
--(axis cs:1,42.4668925301509);

\path [draw=black, semithick]
(axis cs:2,36.0666395562793)
--(axis cs:2,44.7322024683);

\path [draw=black, semithick]
(axis cs:3,61.0175042008297)
--(axis cs:3,76.3089269440621);

\path [draw=black, semithick]
(axis cs:4,69.3433206744441)
--(axis cs:4,76.9361444817626);

\path [draw=black, semithick]
(axis cs:5,55.5946876093886)
--(axis cs:5,73.8681530522232);

\path [draw=black, semithick]
(axis cs:6,52.0674285897514)
--(axis cs:6,58.8340715954086);

\path [draw=black, semithick]
(axis cs:7,40.5224158121861)
--(axis cs:7,60.303745168883);

\path [draw=black, semithick]
(axis cs:8,43.6981218410896)
--(axis cs:8,61.0658979549776);

\path [draw=black, semithick]
(axis cs:9,30.6546384700659)
--(axis cs:9,63.0348062858274);

\path [draw=black, semithick]
(axis cs:10,7.04106914382069)
--(axis cs:10,61.5165545255605);

\path [draw=black, semithick]
(axis cs:11,3.74147457076407)
--(axis cs:11,41.5615959567814);

\path [draw=black, semithick]
(axis cs:12,64.5352709530321)
--(axis cs:12,71.2678721719842);

\path [draw=black, semithick]
(axis cs:13,58.4626942335717)
--(axis cs:13,67.1017225728624);

\path [draw=black, semithick]
(axis cs:14,64.9956484006817)
--(axis cs:14,76.7535618100769);

\path [draw=black, semithick]
(axis cs:15,51.0163833017411)
--(axis cs:15,74.4801044725732);

\path [draw=black, semithick]
(axis cs:16,47.9048902755451)
--(axis cs:16,72.7876768066756);

\path [draw=black, semithick]
(axis cs:17,-1.35286064644973)
--(axis cs:17,33.0381944307293);

\addplot [semithick, black, mark=-, mark size=3, mark options={solid}, only marks, forget plot]
table {%
0 47.0319880887532
1 33.5834137395823
2 36.0666395562793
3 61.0175042008297
4 69.3433206744441
5 55.5946876093886
6 52.0674285897514
7 40.5224158121861
8 43.6981218410896
9 30.6546384700659
10 7.04106914382069
11 3.74147457076407
12 64.5352709530321
13 58.4626942335717
14 64.9956484006817
15 51.0163833017411
16 47.9048902755451
17 -1.35286064644973
};
\addplot [semithick, black, mark=-, mark size=3, mark options={solid}, only marks, forget plot]
table {%
0 50.8560676949342
1 42.4668925301509
2 44.7322024683
3 76.3089269440621
4 76.9361444817626
5 73.8681530522232
6 58.8340715954086
7 60.303745168883
8 61.0658979549776
9 63.0348062858274
10 61.5165545255605
11 41.5615959567814
12 71.2678721719842
13 67.1017225728624
14 76.7535618100769
15 74.4801044725732
16 72.7876768066756
17 33.0381944307293
};

\addplot [semithick, blue, dashed]
table {%
-1 48.9440278918437
18 48.9440278918437
};
\addlegendentry{No Comm., Neighbor Visible, LIDAR on}
\addplot [semithick, red, dashed]
table {%
-1 38.5251531348666
18 38.5251531348666
};
\addlegendentry{No Comm., Neighbor Invisible, LIDAR off}

\end{axis}
\end{tikzpicture}
    \caption{Returns of the 3-agent Multi-Walker with various communication protocol.}
    \label{fig:multi_walker_results}
\end{figure}

The results are shown in \cref{fig:multi_walker_results} for $3$ agents averaged over 5 random seeds. We ran experiments for various communication protocols. In particular, we ran experiments with no communication \textbf{with} neighbor visibility and LIDAR turned on (original observation configuration, labeled with \texttt{no\_T}), and also no communication \textbf{without} neighbor visibility and LIDAR (observation configuration with scarcest info, labeled with \texttt{no\_F}).

The results show observe that communicative methods \emph{outperform} the original observation configuration \texttt{no\_T}. This can be explained by the fact that \texttt{no\_T} only has neighbor information available for each individual agent, while communicative methods can pass the information of non-neighboring agents to each other, providing a wider perception field. Continuous messaging with low bandwidth (\texttt{c4}) has poor performance. Continuous messaging with larger bandwidth ($\ge 8$) has much better performance, within which \texttt{c16} performs best. Bit-string messaging performs worse than continuous messaging and one-hot messaging. In contrast with Pulling Levers and Predator-Prey, the performance of bit-string messaging deteriorates when the bandwidth or vocabulary size increases. And the variance in performance also greatly increases accordingly. One-hot messaging has similar performance as continuous messaging in Multi-Walker, among which \texttt{o32} performs the best. The performance dramatically drops (below \texttt{no\_F}) as the bandwidth increases to 256. Similar to Pulling Levers, overly large bandwidths are harmful for learning emergent communication protocols.

\subsection{Ablation with Self-Attention}
\begin{figure}[t]
\begin{tikzpicture}

\definecolor{color0}{rgb}{0.86,0.3712,0.34}
\definecolor{color1}{rgb}{0.86,0.594057142857143,0.34}
\definecolor{color2}{rgb}{0.86,0.816914285714286,0.34}
\definecolor{color3}{rgb}{0.680228571428571,0.86,0.34}
\definecolor{color4}{rgb}{0.457371428571428,0.86,0.34}
\definecolor{color5}{rgb}{0.34,0.86,0.445485714285714}
\definecolor{color6}{rgb}{0.34,0.86,0.668342857142857}
\definecolor{color7}{rgb}{0.34,0.8288,0.86}
\definecolor{color8}{rgb}{0.34,0.605942857142857,0.86}
\definecolor{color9}{rgb}{0.34,0.383085714285714,0.86}
\definecolor{color10}{rgb}{0.519771428571428,0.34,0.86}
\definecolor{color11}{rgb}{0.742628571428571,0.34,0.86}
\definecolor{color12}{rgb}{0.86,0.34,0.754514285714286}
\definecolor{color13}{rgb}{0.86,0.34,0.531657142857143}

\node[scale=0.65] () at (0, -0.47) {$|\mathcal{V}|$};

\begin{axis}[
legend cell align={left},
legend style={fill opacity=0.8, draw opacity=1, text opacity=1, draw=white!80!black},
tick align=outside,
tick pos=left,
x grid style={white!69.0196078431373!black},
xlabel={Communication Protocols},
xmin=-1, xmax=14,
xtick style={color=black},
xtick={0,1,2,3,4,5,6,7,8,9,10,11,12,13},
xticklabels={no\_T\\0,no\_F\\0,c8\\$\infty$,c8ua\\$\infty$,c32\\$\infty$,c32ua\\$\infty$,b6\\$64$,b6ua\\$64$,b16\\$2^{16}$,b16ua\\$2^{16}$,o32\\$32$,o32ua\\$32$,o64\\$64$,o64ua\\$64$},
xticklabel style={align=center},
y grid style={white!69.0196078431373!black},
ylabel={Average Return},
ymin=28.5, ymax=35.5,
ytick style={color=black},
every tick label/.append style={scale=0.53},
every axis label/.append style={scale=0.8},
legend style={nodes={scale=0.7, transform shape}},
width=3.38in,
height=2.2in,
]
\draw[draw=none,fill=color0] (axis cs:-0.4,0) rectangle (axis cs:0.4,34.0665469625711);
\draw[draw=none,fill=color1] (axis cs:0.6,0) rectangle (axis cs:1.4,29.9712101970206);
\draw[draw=none,fill=color2] (axis cs:1.6,0) rectangle (axis cs:2.4,32.6842143898773);
\draw[draw=none,fill=color2,postaction={pattern=north east lines}] (axis cs:2.6,0) rectangle (axis cs:3.4,32.5228332994493);
\draw[draw=none,fill=color4] (axis cs:3.6,0) rectangle (axis cs:4.4,32.9087645384683);
\draw[draw=none,fill=color4,postaction={pattern=north east lines}] (axis cs:4.6,0) rectangle (axis cs:5.4,32.7550171385799);
\draw[draw=none,fill=color6] (axis cs:5.6,0) rectangle (axis cs:6.4,31.4665703988891);
\draw[draw=none,fill=color6,postaction={pattern=north east lines}] (axis cs:6.6,0) rectangle (axis cs:7.4,30.7836276308523);
\draw[draw=none,fill=color8] (axis cs:7.6,0) rectangle (axis cs:8.4,32.024841266935);
\draw[draw=none,fill=color8,postaction={pattern=north east lines}] (axis cs:8.6,0) rectangle (axis cs:9.4,31.0697581964996);
\draw[draw=none,fill=color10] (axis cs:9.6,0) rectangle (axis cs:10.4,31.6675999017907);
\draw[draw=none,fill=color10,postaction={pattern=north east lines}] (axis cs:10.6,0) rectangle (axis cs:11.4,31.0594281651763);
\draw[draw=none,fill=color12] (axis cs:11.6,0) rectangle (axis cs:12.4,31.804926554153);
\draw[draw=none,fill=color12,postaction={pattern=north east lines}] (axis cs:12.6,0) rectangle (axis cs:13.4,31.2244865563474);
\path [draw=black, semithick]
(axis cs:0,33.9164919924203)
--(axis cs:0,34.2166019327218);

\path [draw=black, semithick]
(axis cs:1,29.7654508958223)
--(axis cs:1,30.1769694982188);

\path [draw=black, semithick]
(axis cs:2,32.5411002019049)
--(axis cs:2,32.8273285778497);

\path [draw=black, semithick]
(axis cs:3,32.4718618100222)
--(axis cs:3,32.5738047888765);

\path [draw=black, semithick]
(axis cs:4,32.684540321747)
--(axis cs:4,33.1329887551896);

\path [draw=black, semithick]
(axis cs:5,32.5694328334056)
--(axis cs:5,32.9406014437541);

\path [draw=black, semithick]
(axis cs:6,31.3988955603834)
--(axis cs:6,31.5342452373948);

\path [draw=black, semithick]
(axis cs:7,30.5472387992635)
--(axis cs:7,31.020016462441);

\path [draw=black, semithick]
(axis cs:8,31.9083466339571)
--(axis cs:8,32.1413358999129);

\path [draw=black, semithick]
(axis cs:9,30.8389117522902)
--(axis cs:9,31.3006046407091);

\path [draw=black, semithick]
(axis cs:10,31.5140022572629)
--(axis cs:10,31.8211975463185);

\path [draw=black, semithick]
(axis cs:11,30.9145780562242)
--(axis cs:11,31.2042782741283);

\path [draw=black, semithick]
(axis cs:12,31.7041678389236)
--(axis cs:12,31.9056852693824);

\path [draw=black, semithick]
(axis cs:13,31.0137891520755)
--(axis cs:13,31.4351839606193);

\addplot [semithick, black, mark=-, mark size=3, mark options={solid}, only marks, forget plot]
table {%
0 33.9164919924203
};
\addplot [semithick, black, mark=-, mark size=3, mark options={solid}, only marks, forget plot]
table {%
0 34.2166019327218
};
\addplot [semithick, black, mark=-, mark size=3, mark options={solid}, only marks, forget plot]
table {%
1 29.7654508958223
};
\addplot [semithick, black, mark=-, mark size=3, mark options={solid}, only marks, forget plot]
table {%
1 30.1769694982188
};
\addplot [semithick, black, mark=-, mark size=3, mark options={solid}, only marks, forget plot]
table {%
2 32.5411002019049
};
\addplot [semithick, black, mark=-, mark size=3, mark options={solid}, only marks, forget plot]
table {%
2 32.8273285778497
};
\addplot [semithick, black, mark=-, mark size=3, mark options={solid}, only marks, forget plot]
table {%
3 32.4718618100222
};
\addplot [semithick, black, mark=-, mark size=3, mark options={solid}, only marks, forget plot]
table {%
3 32.5738047888765
};
\addplot [semithick, black, mark=-, mark size=3, mark options={solid}, only marks, forget plot]
table {%
4 32.684540321747
};
\addplot [semithick, black, mark=-, mark size=3, mark options={solid}, only marks, forget plot]
table {%
4 33.1329887551896
};
\addplot [semithick, black, mark=-, mark size=3, mark options={solid}, only marks, forget plot]
table {%
5 32.5694328334056
};
\addplot [semithick, black, mark=-, mark size=3, mark options={solid}, only marks, forget plot]
table {%
5 32.9406014437541
};
\addplot [semithick, black, mark=-, mark size=3, mark options={solid}, only marks, forget plot]
table {%
6 31.3988955603834
};
\addplot [semithick, black, mark=-, mark size=3, mark options={solid}, only marks, forget plot]
table {%
6 31.5342452373948
};
\addplot [semithick, black, mark=-, mark size=3, mark options={solid}, only marks, forget plot]
table {%
7 30.5472387992635
};
\addplot [semithick, black, mark=-, mark size=3, mark options={solid}, only marks, forget plot]
table {%
7 31.020016462441
};
\addplot [semithick, black, mark=-, mark size=3, mark options={solid}, only marks, forget plot]
table {%
8 31.9083466339571
};
\addplot [semithick, black, mark=-, mark size=3, mark options={solid}, only marks, forget plot]
table {%
8 32.1413358999129
};
\addplot [semithick, black, mark=-, mark size=3, mark options={solid}, only marks, forget plot]
table {%
9 30.8389117522902
};
\addplot [semithick, black, mark=-, mark size=3, mark options={solid}, only marks, forget plot]
table {%
9 31.3006046407091
};
\addplot [semithick, black, mark=-, mark size=3, mark options={solid}, only marks, forget plot]
table {%
10 31.5140022572629
};
\addplot [semithick, black, mark=-, mark size=3, mark options={solid}, only marks, forget plot]
table {%
10 31.8211975463185
};
\addplot [semithick, black, mark=-, mark size=3, mark options={solid}, only marks, forget plot]
table {%
11 30.9145780562242
};
\addplot [semithick, black, mark=-, mark size=3, mark options={solid}, only marks, forget plot]
table {%
11 31.2042782741283
};
\addplot [semithick, black, mark=-, mark size=3, mark options={solid}, only marks, forget plot]
table {%
12 31.7041678389236
};
\addplot [semithick, black, mark=-, mark size=3, mark options={solid}, only marks, forget plot]
table {%
12 31.9056852693824
};
\addplot [semithick, black, mark=-, mark size=3, mark options={solid}, only marks, forget plot]
table {%
13 31.0137891520755
};
\addplot [semithick, black, mark=-, mark size=3, mark options={solid}, only marks, forget plot]
table {%
13 31.4351839606193
};
\addplot [semithick, blue, dashed]
table {%
-1 34.06
15 34.06
};
\addlegendentry{No Comm., Agent Visible}
\addplot [semithick, red, dashed]
table {%
-1 29.97
15 29.97
};
\addlegendentry{No Comm., Agent Invisible}
\end{axis}

\end{tikzpicture}
    \caption{Average return comparison between self-attention and uniform attention from the 4-agent Predator-Prey. Uniform attention bars are shaded and labeled with \texttt{ua} suffix.}
    \label{fig:ua}
\end{figure}
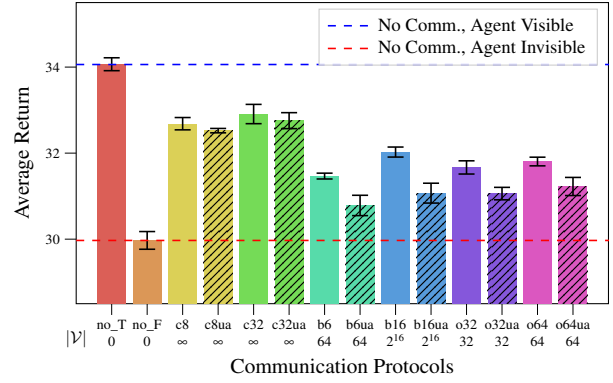

In this analysis, we try to see the contribution of the self-attention to message aggregation. We run self-attention ablation experiments in the 4-agent Predator-Prey environment (the same as \cref{sec:predprey_results}). In \cref{fig:ua}, we show the performance comparison of communication protocols with self-attention (non-shaded) and their counterparts with uniform attention (shaded and with the \texttt{ua} suffix). The results are averaged over 5 random seeds. Uniform attention means that we circumvent the multiplicative attention operation and assign equal attention weight $1/n$ to each agent. 

We observe that uniform attention poses a much more significant negative performance impact on the discrete messaging than on the continuous messaging. Possible explanations are that continuous messages can make up for the difference in attention weights by messages themselves, but different bit-string messages can have drastically different meanings and their limited representational power cannot replace the contributions of learned self-attention weights.

\subsection{Positive Listening and Signaling of Communication}
We design supervised learning tasks~\citep{li2020deep} to measure positive listening and signaling of communication.

\subsubsection{Positive Listening}
Positive listening quantifies the degree to which received messages are influencing an agent's behaviors ~\citep{lowe2019pitfalls, jaques2019social}.
It can be measured with a supervised learning task by using the masked aggregated message $\bar{m}_{aggr, i}$ \emph{received} by agent $i$ to predict agent $i$'s actions. The masking is done to remove agent $i$'s contribution to the aggregated message.

To compute the masked aggregated message $\bar{m}_{aggr, i}$, we first mask the self-attention weight $w_{ii}$ to zero $w_{ii} = 0$, then re-normalize the attention weights $\bar{w}_{ij} = w_{ij} / \sum_{k=1,\dots,n} w_{ik}$, and finally compute the masked aggregated message using the re-normalized attention weights $\bar{m}_{aggr, i} = \sum_{j=1,\dots,n} \bar{w}_{ij} m_j$. 

We formulate the supervised learn task as $\hat{a}_i = f(\bar{m}_{aggr, i}; \phi)$ with loss $L = \text{CrossEntropy}(\hat{a}_i, a_i)$ \citep{li2020deep}. 
Theoretically, with a finite amount of data, if $m_{aggr, i}$ is more correlated with $a_i$, i.e. if positive listening is strong, the classifier $f(\cdot; \phi)$ can produce a higher action prediction accuracy.  We use a simple multi-layer perceptron (MLP) classifier parameterized by $\phi$, with a single 128-unit hidden layer and ReLU as activation. 

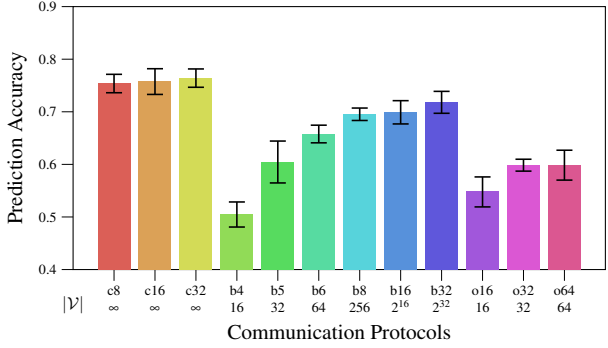
\begin{figure}[t]
\begin{tikzpicture}

\definecolor{color0}{rgb}{0.86,0.3712,0.34}
\definecolor{color1}{rgb}{0.86,0.6312,0.34}
\definecolor{color2}{rgb}{0.8288,0.86,0.34}
\definecolor{color3}{rgb}{0.5688,0.86,0.34}
\definecolor{color4}{rgb}{0.34,0.86,0.3712}
\definecolor{color5}{rgb}{0.34,0.86,0.6312}
\definecolor{color6}{rgb}{0.34,0.8288,0.86}
\definecolor{color7}{rgb}{0.34,0.5688,0.86}
\definecolor{color8}{rgb}{0.3712,0.34,0.86}
\definecolor{color9}{rgb}{0.6312,0.34,0.86}
\definecolor{color10}{rgb}{0.86,0.34,0.8288}
\definecolor{color11}{rgb}{0.86,0.34,0.5688}

\node[scale=0.65] () at (0, -0.47) {$|\mathcal{V}|$};

\begin{axis}[
tick align=outside,
tick pos=left,
x grid style={white!69.0196078431373!black},
xlabel={Communication Protocols},
xmin=-1, xmax=12,
xtick style={color=black},
xtick={0,1,2,3,4,5,6,7,8,9,10,11},
xticklabels={c8\\$\infty$,c16\\$\infty$,c32\\$\infty$,b4\\$16$,b5\\$32$,b6\\$64$,b8\\$256$,b16\\$2^{16}$,b32\\$2^{32}$,o16\\$16$,o32\\$32$,o64\\$64$},
xticklabel style={align=center},
y grid style={white!69.0196078431373!black},
ylabel={Prediction Accuracy},
ymin=0.4, ymax=0.9,
ytick style={color=black},
ytick={0.4,0.5,0.6,0.7,0.8,0.9,1,1.1},
yticklabels={0.4,0.5,0.6,0.7,0.8,0.9,1.0,1.1},
every tick label/.append style={scale=0.53},
every axis label/.append style={scale=0.8},
width=3.38in,
height=2in,
]
\draw[draw=none,fill=color0] (axis cs:-0.4,0) rectangle (axis cs:0.4,0.7537265625);
\draw[draw=none,fill=color1] (axis cs:0.6,0) rectangle (axis cs:1.4,0.757453125);
\draw[draw=none,fill=color2] (axis cs:1.6,0) rectangle (axis cs:2.4,0.7640390625);
\draw[draw=none,fill=color3] (axis cs:2.6,0) rectangle (axis cs:3.4,0.5046953125);
\draw[draw=none,fill=color4] (axis cs:3.6,0) rectangle (axis cs:4.4,0.6045234375);
\draw[draw=none,fill=color5] (axis cs:4.6,0) rectangle (axis cs:5.4,0.65775390625);
\draw[draw=none,fill=color6] (axis cs:5.6,0) rectangle (axis cs:6.4,0.695296875);
\draw[draw=none,fill=color7] (axis cs:6.6,0) rectangle (axis cs:7.4,0.6990234375);
\draw[draw=none,fill=color8] (axis cs:7.6,0) rectangle (axis cs:8.4,0.71798828125);
\draw[draw=none,fill=color9] (axis cs:8.6,0) rectangle (axis cs:9.4,0.5476328125);
\draw[draw=none,fill=color10] (axis cs:9.6,0) rectangle (axis cs:10.4,0.5984765625);
\draw[draw=none,fill=color11] (axis cs:10.6,0) rectangle (axis cs:11.4,0.5985078125);
\path [draw=black, semithick]
(axis cs:0,0.73621875)
--(axis cs:0,0.771234374999999);

\path [draw=black, semithick]
(axis cs:1,0.733007812499999)
--(axis cs:1,0.7818984375);

\path [draw=black, semithick]
(axis cs:2,0.7466140625)
--(axis cs:2,0.7814640625);

\path [draw=black, semithick]
(axis cs:3,0.4808625)
--(axis cs:3,0.528528125);

\path [draw=black, semithick]
(axis cs:4,0.56470078125)
--(axis cs:4,0.64434609375);

\path [draw=black, semithick]
(axis cs:5,0.64097265625)
--(axis cs:5,0.67453515625);

\path [draw=black, semithick]
(axis cs:6,0.68347734375)
--(axis cs:6,0.70711640625);

\path [draw=black, semithick]
(axis cs:7,0.67689921875)
--(axis cs:7,0.72114765625);

\path [draw=black, semithick]
(axis cs:8,0.6971421875)
--(axis cs:8,0.738834375);

\path [draw=black, semithick]
(axis cs:9,0.51912109375)
--(axis cs:9,0.57614453125);

\path [draw=black, semithick]
(axis cs:10,0.58716640625)
--(axis cs:10,0.60978671875);

\path [draw=black, semithick]
(axis cs:11,0.5700890625)
--(axis cs:11,0.6269265625);

\addplot [semithick, black, mark=-, mark size=3, mark options={solid}, only marks]
table {%
0 0.73621875
};
\addplot [semithick, black, mark=-, mark size=3, mark options={solid}, only marks]
table {%
0 0.771234374999999
};
\addplot [semithick, black, mark=-, mark size=3, mark options={solid}, only marks]
table {%
1 0.733007812499999
};
\addplot [semithick, black, mark=-, mark size=3, mark options={solid}, only marks]
table {%
1 0.7818984375
};
\addplot [semithick, black, mark=-, mark size=3, mark options={solid}, only marks]
table {%
2 0.7466140625
};
\addplot [semithick, black, mark=-, mark size=3, mark options={solid}, only marks]
table {%
2 0.7814640625
};
\addplot [semithick, black, mark=-, mark size=3, mark options={solid}, only marks]
table {%
3 0.4808625
};
\addplot [semithick, black, mark=-, mark size=3, mark options={solid}, only marks]
table {%
3 0.528528125
};
\addplot [semithick, black, mark=-, mark size=3, mark options={solid}, only marks]
table {%
4 0.56470078125
};
\addplot [semithick, black, mark=-, mark size=3, mark options={solid}, only marks]
table {%
4 0.64434609375
};
\addplot [semithick, black, mark=-, mark size=3, mark options={solid}, only marks]
table {%
5 0.64097265625
};
\addplot [semithick, black, mark=-, mark size=3, mark options={solid}, only marks]
table {%
5 0.67453515625
};
\addplot [semithick, black, mark=-, mark size=3, mark options={solid}, only marks]
table {%
6 0.68347734375
};
\addplot [semithick, black, mark=-, mark size=3, mark options={solid}, only marks]
table {%
6 0.70711640625
};
\addplot [semithick, black, mark=-, mark size=3, mark options={solid}, only marks]
table {%
7 0.67689921875
};
\addplot [semithick, black, mark=-, mark size=3, mark options={solid}, only marks]
table {%
7 0.72114765625
};
\addplot [semithick, black, mark=-, mark size=3, mark options={solid}, only marks]
table {%
8 0.6971421875
};
\addplot [semithick, black, mark=-, mark size=3, mark options={solid}, only marks]
table {%
8 0.738834375
};
\addplot [semithick, black, mark=-, mark size=3, mark options={solid}, only marks]
table {%
9 0.51912109375
};
\addplot [semithick, black, mark=-, mark size=3, mark options={solid}, only marks]
table {%
9 0.57614453125
};
\addplot [semithick, black, mark=-, mark size=3, mark options={solid}, only marks]
table {%
10 0.58716640625
};
\addplot [semithick, black, mark=-, mark size=3, mark options={solid}, only marks]
table {%
10 0.60978671875
};
\addplot [semithick, black, mark=-, mark size=3, mark options={solid}, only marks]
table {%
11 0.5700890625
};
\addplot [semithick, black, mark=-, mark size=3, mark options={solid}, only marks]
table {%
11 0.6269265625
};
\end{axis}

\end{tikzpicture}
    \caption{Positive listening analysis on the 4-agent Predator-Prey environment: predicting agent $i$'s actions $a_i$ using the masked aggregated messages $\bar{m}_{aggr,i}$ received by agent $i$. Data are collected from evaluation trajectories of five random runs.}
    \label{fig:pos_lis}
\end{figure}

Results of five random runs from Predator-Prey are shown in \cref{fig:pos_lis}. Environment configuration is the same as that in \cref{sec:predprey_results}.
In general, higher bandwidths or larger vocabulary size can bring a stronger positive listening for all communication protocols. In terms of communication bandwidth, continuous messaging shows the best positive listening, closely followed by bit-string messaging. Bit-string messaging exhibits potential to yield positive listening on par with continuous messaging.
In terms of the vocabulary size instead of the bandwidth, one-hot messaging has a higher action prediction accuracy than bit-string messaging at low vocabulary sizes ($|\mathcal{V}| \le 32$).

A high action prediction accuracy is achieved because the broadcast-and-listen architecture allows individual agents to learn other agents' behaviors and intentions through messages they sent. 

\subsubsection{Positive Signaling}
Positive signaling quantifies the degree to which an agent's sent messages are related to its observations or actions~\citep{lowe2019pitfalls, jaques2019social}.
It can be measured with a supervised learning task by using the message $m_i$ \emph{sent} by agent $i$ to predict agent $i$'s actions. 

Similar to positive listening, we formulate the supervised learn task as $\hat{a}_i = f(m_i; \phi)$ with the same cross entropy loss and network architecture as used for positive listening.
With a finite amount of data, if $m_i$ is more correlated with $a_i$, i.e. if positive signaling is strong, the classifier $f(\cdot; \phi)$ can produce a higher action prediction accuracy. 

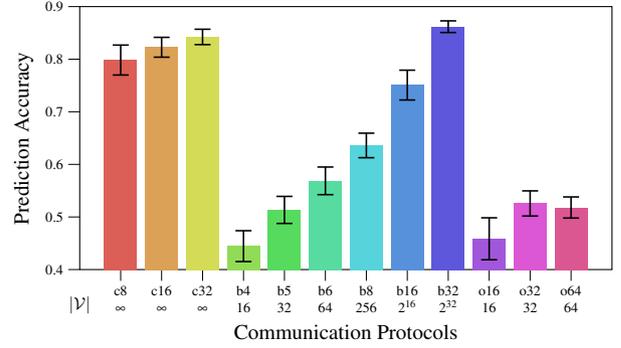
\begin{figure}[t]
\begin{tikzpicture}

\definecolor{color0}{rgb}{0.86,0.3712,0.34}
\definecolor{color1}{rgb}{0.86,0.6312,0.34}
\definecolor{color2}{rgb}{0.8288,0.86,0.34}
\definecolor{color3}{rgb}{0.5688,0.86,0.34}
\definecolor{color4}{rgb}{0.34,0.86,0.3712}
\definecolor{color5}{rgb}{0.34,0.86,0.6312}
\definecolor{color6}{rgb}{0.34,0.8288,0.86}
\definecolor{color7}{rgb}{0.34,0.5688,0.86}
\definecolor{color8}{rgb}{0.3712,0.34,0.86}
\definecolor{color9}{rgb}{0.6312,0.34,0.86}
\definecolor{color10}{rgb}{0.86,0.34,0.8288}
\definecolor{color11}{rgb}{0.86,0.34,0.5688}

\node[scale=0.65] () at (0, -0.47) {$|\mathcal{V}|$};

\begin{axis}[
tick align=outside,
tick pos=left,
x grid style={white!69.0196078431373!black},
xlabel={Communication Protocols},
xmin=-1, xmax=12,
xtick style={color=black},
xtick={0,1,2,3,4,5,6,7,8,9,10,11},
xticklabels={c8\\$\infty$,c16\\$\infty$,c32\\$\infty$,b4\\$16$,b5\\$32$,b6\\$64$,b8\\$256$,b16\\$2^{16}$,b32\\$2^{32}$,o16\\$16$,o32\\$32$,o64\\$64$},
xticklabel style={align=center},
y grid style={white!69.0196078431373!black},
ylabel={Prediction Accuracy},
ymin=0.4, ymax=0.9,
ytick style={color=black},
ytick={0.4,0.5,0.6,0.7,0.8,0.9,1,1.1},
yticklabels={0.4,0.5,0.6,0.7,0.8,0.9,1.0,1.1},
every tick label/.append style={scale=0.53},
every axis label/.append style={scale=0.8},
width=3.38in,
height=2in,
]
\draw[draw=none,fill=color0] (axis cs:-0.4,0) rectangle (axis cs:0.4,0.7983671875);
\draw[draw=none,fill=color1] (axis cs:0.6,0) rectangle (axis cs:1.4,0.8226953125);
\draw[draw=none,fill=color2] (axis cs:1.6,0) rectangle (axis cs:2.4,0.84233203125);
\draw[draw=none,fill=color3] (axis cs:2.6,0) rectangle (axis cs:3.4,0.44452734375);
\draw[draw=none,fill=color4] (axis cs:3.6,0) rectangle (axis cs:4.4,0.51339453125);
\draw[draw=none,fill=color5] (axis cs:4.6,0) rectangle (axis cs:5.4,0.56870703125);
\draw[draw=none,fill=color6] (axis cs:5.6,0) rectangle (axis cs:6.4,0.6360078125);
\draw[draw=none,fill=color7] (axis cs:6.6,0) rectangle (axis cs:7.4,0.750765625);
\draw[draw=none,fill=color8] (axis cs:7.6,0) rectangle (axis cs:8.4,0.8617890625);
\draw[draw=none,fill=color9] (axis cs:8.6,0) rectangle (axis cs:9.4,0.45851953125);
\draw[draw=none,fill=color10] (axis cs:9.6,0) rectangle (axis cs:10.4,0.5256953125);
\draw[draw=none,fill=color11] (axis cs:10.6,0) rectangle (axis cs:11.4,0.518015625);
\path [draw=black, semithick]
(axis cs:0,0.76990078125)
--(axis cs:0,0.82683359375);

\path [draw=black, semithick]
(axis cs:1,0.803825781249999)
--(axis cs:1,0.84156484375);

\path [draw=black, semithick]
(axis cs:2,0.8276984375)
--(axis cs:2,0.856965625);

\path [draw=black, semithick]
(axis cs:3,0.41519609375)
--(axis cs:3,0.47385859375);

\path [draw=black, semithick]
(axis cs:4,0.48766640625)
--(axis cs:4,0.53912265625);

\path [draw=black, semithick]
(axis cs:5,0.5424390625)
--(axis cs:5,0.594975);

\path [draw=black, semithick]
(axis cs:6,0.6126609375)
--(axis cs:6,0.6593546875);

\path [draw=black, semithick]
(axis cs:7,0.722402343749999)
--(axis cs:7,0.77912890625);

\path [draw=black, semithick]
(axis cs:8,0.85070703125)
--(axis cs:8,0.87287109375);

\path [draw=black, semithick]
(axis cs:9,0.41863046875)
--(axis cs:9,0.49840859375);

\path [draw=black, semithick]
(axis cs:10,0.5017828125)
--(axis cs:10,0.5496078125);

\path [draw=black, semithick]
(axis cs:11,0.49794296875)
--(axis cs:11,0.53808828125);

\addplot [semithick, black, mark=-, mark size=3, mark options={solid}, only marks]
table {%
0 0.76990078125
};
\addplot [semithick, black, mark=-, mark size=3, mark options={solid}, only marks]
table {%
0 0.82683359375
};
\addplot [semithick, black, mark=-, mark size=3, mark options={solid}, only marks]
table {%
1 0.803825781249999
};
\addplot [semithick, black, mark=-, mark size=3, mark options={solid}, only marks]
table {%
1 0.84156484375
};
\addplot [semithick, black, mark=-, mark size=3, mark options={solid}, only marks]
table {%
2 0.8276984375
};
\addplot [semithick, black, mark=-, mark size=3, mark options={solid}, only marks]
table {%
2 0.856965625
};
\addplot [semithick, black, mark=-, mark size=3, mark options={solid}, only marks]
table {%
3 0.41519609375
};
\addplot [semithick, black, mark=-, mark size=3, mark options={solid}, only marks]
table {%
3 0.47385859375
};
\addplot [semithick, black, mark=-, mark size=3, mark options={solid}, only marks]
table {%
4 0.48766640625
};
\addplot [semithick, black, mark=-, mark size=3, mark options={solid}, only marks]
table {%
4 0.53912265625
};
\addplot [semithick, black, mark=-, mark size=3, mark options={solid}, only marks]
table {%
5 0.5424390625
};
\addplot [semithick, black, mark=-, mark size=3, mark options={solid}, only marks]
table {%
5 0.594975
};
\addplot [semithick, black, mark=-, mark size=3, mark options={solid}, only marks]
table {%
6 0.6126609375
};
\addplot [semithick, black, mark=-, mark size=3, mark options={solid}, only marks]
table {%
6 0.6593546875
};
\addplot [semithick, black, mark=-, mark size=3, mark options={solid}, only marks]
table {%
7 0.722402343749999
};
\addplot [semithick, black, mark=-, mark size=3, mark options={solid}, only marks]
table {%
7 0.77912890625
};
\addplot [semithick, black, mark=-, mark size=3, mark options={solid}, only marks]
table {%
8 0.85070703125
};
\addplot [semithick, black, mark=-, mark size=3, mark options={solid}, only marks]
table {%
8 0.87287109375
};
\addplot [semithick, black, mark=-, mark size=3, mark options={solid}, only marks]
table {%
9 0.41863046875
};
\addplot [semithick, black, mark=-, mark size=3, mark options={solid}, only marks]
table {%
9 0.49840859375
};
\addplot [semithick, black, mark=-, mark size=3, mark options={solid}, only marks]
table {%
10 0.5017828125
};
\addplot [semithick, black, mark=-, mark size=3, mark options={solid}, only marks]
table {%
10 0.5496078125
};
\addplot [semithick, black, mark=-, mark size=3, mark options={solid}, only marks]
table {%
11 0.49794296875
};
\addplot [semithick, black, mark=-, mark size=3, mark options={solid}, only marks]
table {%
11 0.53808828125
};
\end{axis}

\end{tikzpicture}
    \caption{Positive signaling analysis on the 4-agent Predator-Prey environment: predicting agent $i$'s actions $a_i$ using the message $m_i$ sent by agent $i$. Data are collected from evaluation trajectories of five random runs.}
    \label{fig:pos_sig}
\end{figure}

Results of five random runs from Predator-Prey are shown in \cref{fig:pos_sig}. Continuous messaging shows the strongest positive signaling, while one-hot messaging shows the weakest positive signaling. 
Bit-string messaging shows large gaps between continuous messaging when communication bandwidth is small. The gap decreases as the bandwidth increases. Bit-string messaging outperforms continuous messaging when bandwidth reaches $32$.
One-hot messaging under performs bit-string messaging by a large margin.

\subsection{Human Interpretation and Interaction}

Discrete messaging has a finite vocabulary size that human can easily interpret. We design a protocol to allow a human to interactively send messages to AI agents. 

\begin{figure}[t]
    \centering
    \includegraphics[trim={1cm 0 0 1.5cm}, clip, width=0.42\textwidth]{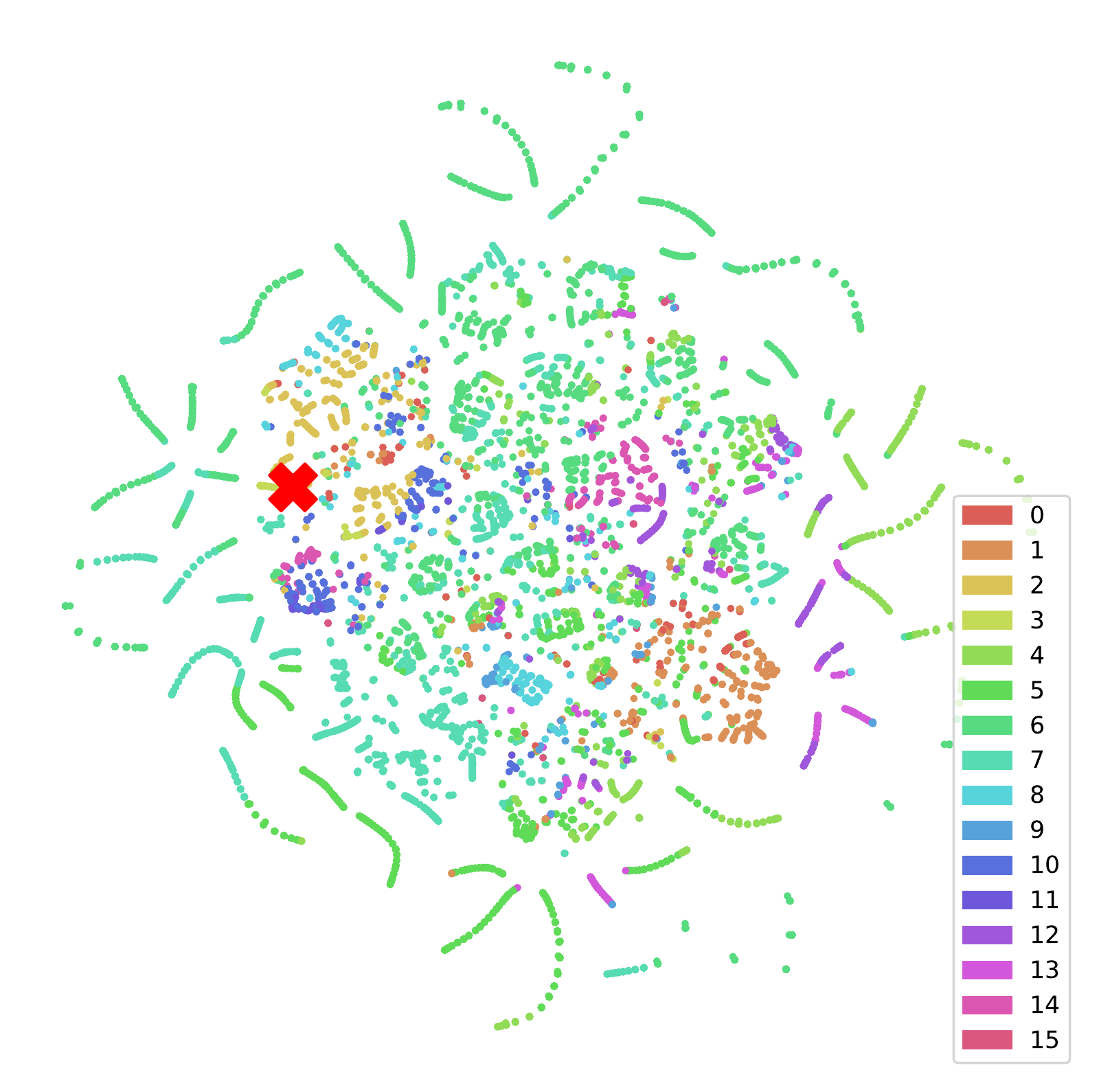}
    \caption{T-SNE clustering of agent observations labeled by their corresponding messages (4-agent Predator-Prey with \texttt{b4}). The red cross represents an example of the projection of a new observation.}
    \label{fig:tsne}
\end{figure}

First, we collect raw observation and message pairs from trained agents. We cluster the raw observations using t-SNE~\citep{van2008visualizing, Policar731877}. Then, we label the raw observations with their corresponding discrete messages. \cref{fig:tsne} shows such clustering and labeling from Predator-Prey with \texttt{b4} (with a vocabulary size of $2^4=16$). Different colors in the clustering plot means different discrete messages. We use this clustering plot as a reference. We observe that the same cluster of raw observations tend to have the same discrete message labels. This provides an entry point for human-agent interaction. We can project new observations into the pre-known reference clusters as shown by the red cross in \cref{fig:tsne}, and empirically select the most probable messages to send to agents. \cref{fig:interaction} shows an outline for the human-agent interaction workflow. 

\begin{figure}[t]
    \centering
\begin{tikzpicture}[>=stealth']
\small
\node[circle, draw=black, line width=0.2mm] (_agent) at (0, 0) {Agent};
\node[circle, draw=black, line width=0.2mm] (_human) at (4, 0) {Human};
\draw [->, black, thick, bend right=25, line width=0.2mm] (_agent) to (_human);
\draw [->, black, thick, bend right=25, line width=0.2mm] (_human) to (_agent);

\node [] () at (2, 1.2) {\begin{tabular}{c}Message selected from\\clustering and projection\end{tabular}};
\node [] () at (2, -1.2) {\begin{tabular}{c}Message selected from\\trained network\end{tabular}};
\end{tikzpicture}
    \caption{Human-agent interaction.}
    \label{fig:interaction}
\end{figure}
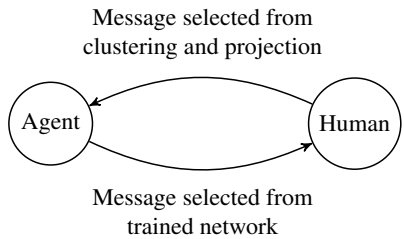

\begin{figure}[t]
    \centering
\begin{tikzpicture}

\definecolor{color0}{rgb}{0.86,0.3712,0.34}
\definecolor{color1}{rgb}{0.86,0.594057142857143,0.34}
\definecolor{color2}{rgb}{0.86,0.816914285714286,0.34}

\begin{axis}[
legend cell align={left},
legend style={fill opacity=0.8, draw opacity=1, text opacity=1, draw=white!80!black},
tick align=outside,
tick pos=left,
x grid style={white!69.0196078431373!black},
xlabel={Message Selection},
xmin=-1, xmax=3,
xtick style={color=black},
xtick={0,1,2},
xticklabels={Agent, Human, Random},
y grid style={white!69.0196078431373!black},
ylabel={Average Return},
ymin=27, ymax=32,
ytick style={color=black},
every tick label/.append style={scale=0.75},
every axis label/.append style={scale=0.85},
legend style={nodes={scale=0.7, transform shape}},
width=3.0in,
height=2.0in,
]
\draw[draw=none,fill=color0] (axis cs:-0.4,0) rectangle (axis cs:0.4,31.3804071635213);
\draw[draw=none,fill=color1] (axis cs:0.6,0) rectangle (axis cs:1.4,30.222);
\draw[draw=none,fill=color2] (axis cs:1.6,0) rectangle (axis cs:2.4,28.33);

\path [draw=black, semithick]
(axis cs:0,31.1379143482224)
--(axis cs:0,31.6228999788202);

\path [draw=black, semithick]
(axis cs:1,29.2654508958223)
--(axis cs:1,31.1769694982188);

\path [draw=black, semithick]
(axis cs:2,27.78)
--(axis cs:2,28.88);

\addplot [semithick, black, mark=-, mark size=3, mark options={solid}, only marks, forget plot]
table {%
0 31.1379143482224
};
\addplot [semithick, black, mark=-, mark size=3, mark options={solid}, only marks, forget plot]
table {%
0 31.6228999788202
};
\addplot [semithick, black, mark=-, mark size=3, mark options={solid}, only marks, forget plot]
table {%
1 29.2654508958223
};
\addplot [semithick, black, mark=-, mark size=3, mark options={solid}, only marks, forget plot]
table {%
1 31.1769694982188
};
\addplot [semithick, black, mark=-, mark size=3, mark options={solid}, only marks, forget plot]
table {%
2 27.78
};
\addplot [semithick, black, mark=-, mark size=3, mark options={solid}, only marks, forget plot]
table {%
2 28.88
};
\end{axis}

\end{tikzpicture}
    \caption{Human-agent interaction results and comparison for Predator-Prey with \texttt{b4}.}
    \label{fig:interaction_results}
\end{figure}
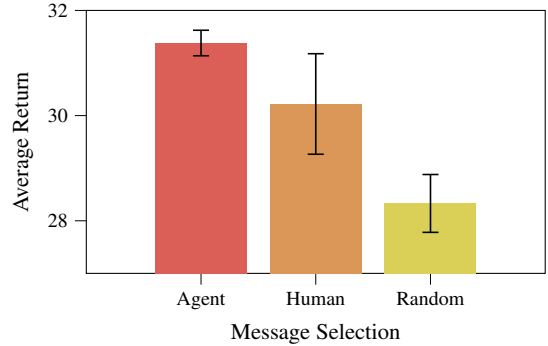

We design an experiment for letting human to select messages for \textbf{one} agent, with the rest of the agents selecting messages using the trained network (actual human-agent interaction experiments have text labels besides colors on the reference clustering points to assist in message selection).
We compare the performance with agent selecting messages and random messages for one agent. 
\cref{fig:interaction_results} shows the results for the 4-agent Predator-Prey with 4-bit-string messaging (vocabulary size $2^4=16$). Environment configuration is the same as that in \cref{sec:predprey_results}. Agent and random selection are averaged over 200 episodes and human selection is averaged over 20 episodes. We can see human message selection has a bit worse performance than agent selection. The performance gap arises from the empirical prediction error when a human selects messages from their estimate using clustering and projection. Human message selection outperforms random selection.

In summary, the human-agent interaction protocol described in this analysis shows the interpretability of discrete messaging and demonstrates a way to integrate both human and AI agents.
\section{Conclusions}
In this work, we present a broadcast-and-listen model that enables end-to-end learning of emergent discrete message communication. We demonstrate that the bandwidth and the
vocabulary size of discrete messaging affects its performance. In some domains, discrete message communication can yield return performance and positive listening and signaling on par with or exceeding continuous message communication. Since discrete messages are easier to interpret by humans, we propose a human-agent interaction protocol that allows human to send discrete messages to agents. For future work, we would like to try multi-headed attention~\citep{vaswani2017attention} for the information aggregation process and new metrics other than average returns to more directly measure communication capabilities~\citep{lowe2019pitfalls}. 

\section{Acknowledgements}
DISTRIBUTION STATEMENT A. Approved for public release. Distribution is unlimited.
This material is based upon work supported by the Under Secretary of Defense for Research and Engineering under Air Force Contract No. FA8702-15-D-0001. Any opinions, findings, conclusions or recommendations expressed in this material are those of the author(s) and do not necessarily reflect the views of the Under Secretary of Defense for Research and Engineering.

\bibliography{main}
\end{document}